%% file: main.tex
\documentclass[10pt,twocolumn,letterpaper]{article}

\usepackage{iccv}              
\usepackage{multirow}  
\usepackage{ragged2e}
\usepackage{nicematrix}
\usepackage{algorithm}
\usepackage{algpseudocode}
\usepackage{graphicx}
\usepackage{soul}
\usepackage{fontawesome5}


\definecolor{iccvblue}{rgb}{0.21,0.49,0.74}
\usepackage[pagebackref,breaklinks,colorlinks,allcolors=iccvblue]{hyperref}

\def\paperID{9070} 
\def\confName{ICCV}
\def\confYear{2025}

\def\algo{GraPE}
\def\algosim{GraPE\includegraphics[width=0.35cm]{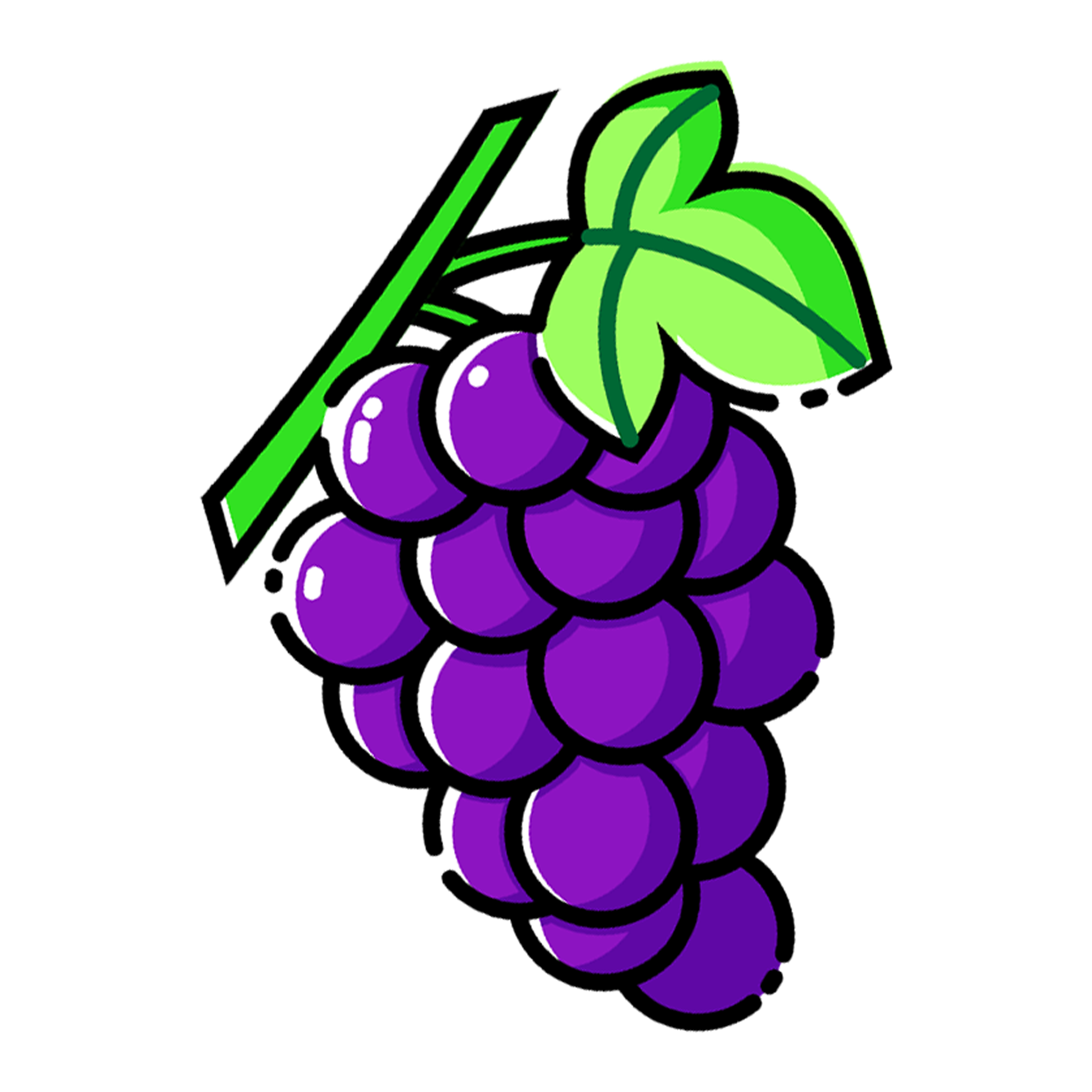}}
\def\algosimtitle{GraPE\includegraphics[width=0.5cm]{assets/grape2.png}}

\newcolumntype{M}[1]{>{\centering\raggedright\arraybackslash}m{#1}}
\newcolumntype{C}[1]{>{\centering\arraybackslash}m{#1}}

\title{\algosimtitle: A \underline{G}ene\underline{ra}te-\underline{P}lan-\underline{E}dit Framework for Compositional T2I Synthesis}

\author{Ashish Goswami\textsuperscript{1},\quad Satyam Kumar Modi\textsuperscript{1*},\quad 
Santhosh Rishi Deshineni\textsuperscript{1*}, \\
Harman Singh\textsuperscript{1}, \quad Prathosh A. P\textsuperscript{2}, \quad Parag Singla\textsuperscript{1}\\
\textsuperscript{1}IIT-Delhi,\quad \textsuperscript{2}IISc Bangalore\\
{\tt\small {ashish.goswami@scai.iitd.ac.in,\quad parags@cse.iitd.ac.in}}\\
}

\begin{document}
\maketitle
\input{sec/0_abstract}    
\input{sec/1_intro}
\input{sec/2_related_works}
\input{sec/3_method_1}

\input{sec/4_experiments}
\input{sec/5_conclusion}
{
    \small
    \bibliographystyle{ieeenat_fullname}
    \bibliography{main}
}

\input{sec/X_suppl}

\end{document}

%% file: sec/0_abstract.tex
\begin{abstract}
\label{sec:abstract}
Text-to-image (T2I) generation has seen significant progress with diffusion models, enabling generation of photo-realistic images from text prompts. Despite this progress, existing methods still face challenges in following complex text prompts, especially those requiring compositional and multi-step reasoning. Given such complex instructions, SOTA models often make mistakes in faithfully modeling object attributes, and relationships among them. 

In this work, we present an alternate paradigm for T2I synthesis, decomposing the task of complex multi-step generation into three steps, (a) {\em Generate}: we first generate an image using existing diffusion models (b) {\em Plan}: we make use of Multi-Modal LLMs (MLLMs) to identify the mistakes in the generated image expressed in terms of individual objects and their properties, and produce a sequence of corrective steps required in the form of an {\em edit-plan}. (c) {\em Edit}: we make use of an existing text-guided image editing models to sequentially execute our edit-plan over the generated image to get the desired image which is faithful to the original instruction. Our approach derives its strength from the fact that it is modular in nature, is training free, and can be applied over any combination of image generation and editing models. As an added contribution, we also develop a model capable of compositional editing, which further helps improve the overall accuracy of our proposed approach. Our method flexibly trades inference time compute with performance on compositional text prompts. We perform extensive experimental evaluation across 3 benchmarks and 13 T2I models including DALLE-3 and the latest -- SD-3.5-Large. Our approach not only improves the performance of the SOTA models, by upto 3 points, it also reduces the performance gap between weaker and stronger models. The code is available at \href{https://dair-iitd.github.io/GraPE/}{https://dair-iitd.github.io/GraPE/}


\begin{figure}
  \centering
  \includegraphics[width=1.0\linewidth]{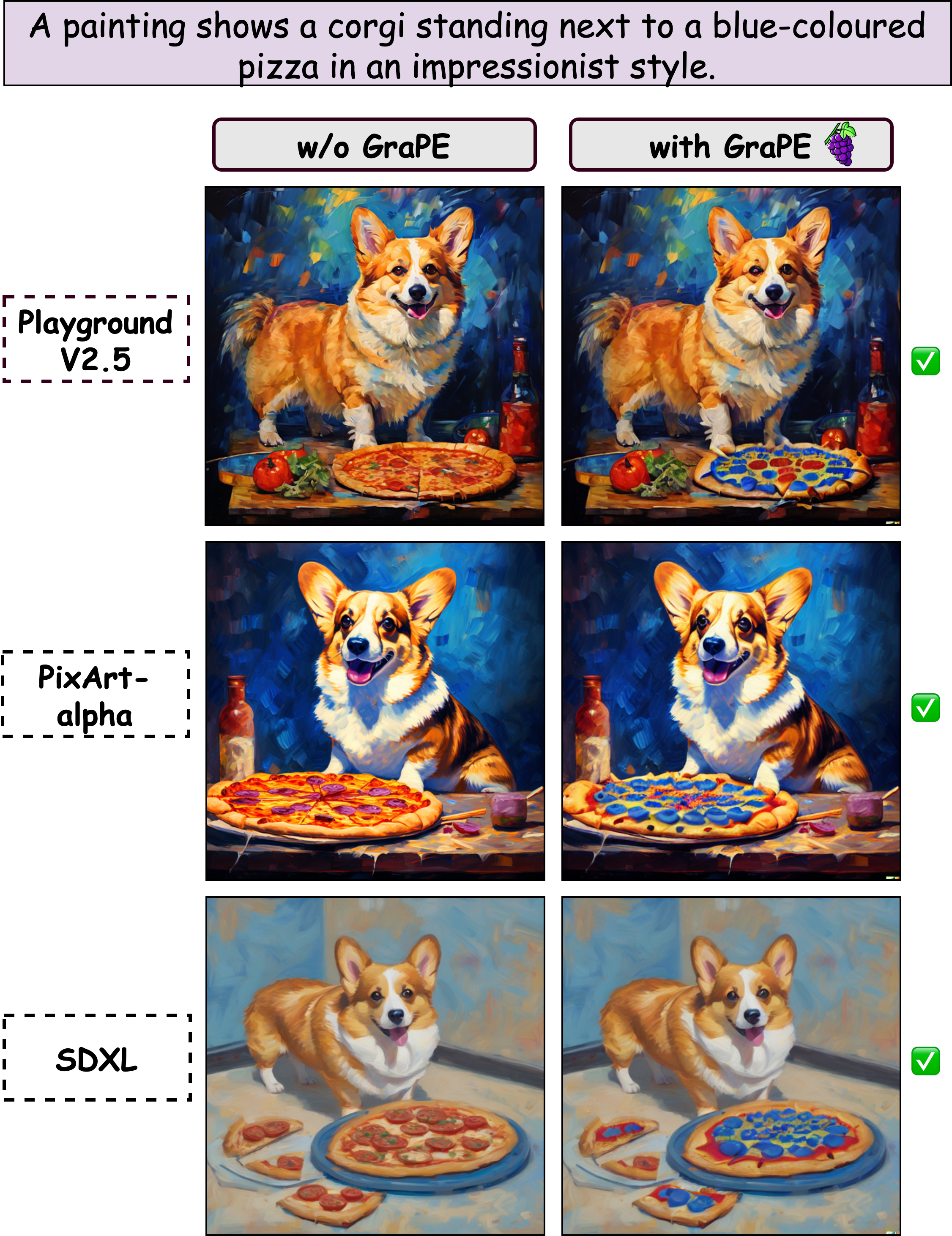}
  \caption{Illustration of \algo's capability to align the image with the input prompt. \textit{Left}: Generations from various state-of-the-art diffusion models~\citep{li2024playgroundv25insightsenhancing, chen2023pixartalphafasttrainingdiffusion, podell2023sdxlimprovinglatentdiffusion}, along with their inaccuracies. \textit{Right}: Images produced by \algo, in a completely training free manner.}
  \label{fig:teaser}
\end{figure}

\end{abstract} 

%% file: sec/1_intro.tex
\section{Introduction}
\label{sec:intro}
Diffusion models~\citep{sohldickstein2015deepunsupervisedlearningusing}, have revolutionized the space of generating photo-realistic images with multiple works showing their superior performance compared to their older competitors such as GANs, VAEs and Flow-based models~\citep{Dhariwal2021DiffusionMB, Saharia2022PhotorealisticTD, Nichol2021ImprovedDD}. Of specific interest has been their ability to generate images from text. Despite the success, it has been observed that SOTA diffusion models still struggle to generate accurate images for instructions which involve multi-step compositional reasoning, often resulting in errors over object attributes as well as their interactions with other objects in the image~\citep{chefer2023attendandexciteattentionbasedsemanticguidance, huang2023t2icompbenchcomprehensivebenchmarkopenworld, ramesh2022hierarchicaltextconditionalimagegeneration}. Figure~\ref{fig:teaser} shows an example of a text-prompt and the corresponding images generated by 3 of the SOTA text-to-image generation models. Several reasons have been pointed for the relative performance on such complex instructions, including lack of appropriate training data, and use of weak text-encoders to name a few. This has naturally limited their applicability for the task of automated and reliable image generation in real life scenarios. To make matters worse, image-editing has been left even further behind, with existing models hardly capable of correctly handing complex edit tasks. 

To overcome these challenges, recent work has seen several credible attempts to address the challenge of generating images faithful to complex instructions. Broadly, these can be divided in two categories (1) those requiring fine-tuning of existing generation models, examples include~\citep{jiang2024comataligningtexttoimagediffusion, hu2024ellaequipdiffusionmodels, wang2024tokencomposetexttoimagediffusiontokenlevel, chatterjee2024gettingrightimprovingspatial, wu2023paragraphtoimagegenerationinformationenricheddiffusion} (2) and those which are based only on adapting the inference procedure, examples include ~\citep{chefer2023attendandexciteattentionbasedsemanticguidance, feng2023trainingfreestructureddiffusionguidance, lian2024llmgroundeddiffusionenhancingprompt, agarwal2023astartesttimeattentionsegregation}. While these approaches have been able to show some improvements over the base diffusion models, the problem remains far from solved. We survey the existing related works in detail in Section~\ref{sec:related}.

In this work, we present a novel approach which is based on the following observations: Instead of performing the image synthesis for a complex compositional instruction in one go, we can redefine the task into iterative refinement of a partially faithful generated image. Our approach can be described as a pipeline of the following 3 steps: (a) {\em Generate:} Initial generation using existing T2I models (b) {\em Plan:} Identification of errors in the form of a series of simple edits (largely over individual objects and their relations) required to fix the original generation; the edits are expressed in the form of a \textit{sequential editing plan} generated via the use of a MLLM (c) {\em Edit} carrying out the edit plan in a sequential manner over the original generation to get the desired image. We refer to our approach as {\em GeneRAte-Plan-Edit} (\algo). To the best of our knowledge, we are the first ones to decompose the task of fine-grained generation as a series of simpler (atomic) edits which are preceded by a partially correct generation.

We perform a series of experiments to evaluate the efficacy of our approach over three benchmark datasets. We compare with 13 different T2I models, and show improvement in each case, with improvements up-to a significant 20+ points in some cases. We also note that we see a more significant improvement in weaker (typically smaller) generation models, resulting in narrowing of performance gap between smaller and larger SoTA models. We also perform ablations showing the value of our object-centric approach for generating edit plans. Along with this, significant boost in performance is shown when we use the compositional image editor that we develop, as part of our pipeline. We also present detailed error analysis and point to limitations of our model in our experimental analysis.

The contributions of our paper can be summarized as follows: (a) We propose a simple yet effective approach for improving the performance of T2I models via a generate-plan-edit pipeline. (b) Our approach is guided by object-centric edit plans, and exploits the power of editing models which can handle compositions well. (c) Our approach is modular, and can be used with any existing base generation or editing model (d) We perform a series of experiments demonstrating the efficacy of our approach over a large number of T2I models, and also present detailed insights into the working of model.





%% file: sec/2_related_works.tex
\section{Related Work}
\label{sec:related}

\paragraph{Compositional Image Generation:} T2I models struggle with compositional image generation, such as generating objects with correct attributes, adhering to object relations, etc. \citep{feng2023trainingfree,bakr2023hrs,gokhale2023benchmarkingspatialrelationshipstexttoimage}. Several works aim to improve the compositional image generation capabilities of T2I diffusion models. Some prior works intervene on the cross-attention maps in diffusion models which are responsible for incorporating information form the text into the image. \citep{chefer2023attendandexciteattentionbasedsemanticguidance, agarwal2023astartesttimeattentionsegregation} strengthen the attention activations of neglected concepts in the image during the generative process using inference time losses, primarily targeting the reduction of catastrophic neglect and incorrect attribute binding. \citep{wang2024tokencomposetexttoimagediffusiontokenlevel,jiang2024comataligningtexttoimagediffusion} proposes training models focused on alignment of Cross-attention maps for each token with pseudo-ground truth segmentation maps. \cite{feng2023trainingfreestructureddiffusionguidance} uses linguistic structures and structured representations of text such as constituency tree, for manipulating cross-attention representations. However, all these works often focus on narrow compositional concepts, such as only attribute binding or reducing catastrophic neglect of objects, than being a general purpose method for aligning images with text. \citep{wang2024genartist} uses MLLMs as an agent with an extensive set of tools for decomposing and planning the generation process into multiple steps. \cite{yang2024mastering} uses an LLM for planning the generation process and suggesting sub-regions for region-wise diffusion, however, it may struggle with either with decomposing into or merging of these regions, since their re-weighting and sampling method is different from what diffusion generative models are trained for. \citep{lian2024llmgroundeddiffusionenhancingprompt,Cho2023VPT2I,feng2024rannitamingtexttoimagediffusion, nie2024BlobGEN} utilize LLMs for generating intermediate  representations such as layouts, panels or blobs followed by generating using specialized models that require such complex annotations to be trained. \citep{gao2024luminat2xtransformingtextmodality, zhuo2024luminanextmakingluminat2xstronger, liu2024playgroundv3improvingtexttoimage,xie2024sana, chen2024pixartsigmaweaktostrongtrainingdiffusion} utilize LLMs such as Gemma, LLaMA , T5-XXL \citep{gemmateam2024gemmaopenmodelsbased, touvron2023llama2openfoundation, raffel2023exploringlimitstransferlearning} as text-encoders for improved text representations compared to contrastively trained models like CLIP \citep{radford2021learningtransferablevisualmodels} that are largely agnostic to word order \cite{yuksekgonul2023visionlanguagemodelsbehavelike, kamath2023textencodersbottleneckcompositionality}. Orthogonally, we use MLLMs for planning the sequential editing process based on the initial generated image, and use general purpose generator and editor models in their native fashion.


\paragraph{Instruction Guided Image Editing:} Editing images with human written instructions have seen significant interest in the community \cite{brooks2023instructpix2pixlearningfollowimage, zhang2024magicbrushmanuallyannotateddataset, zhang2021text, zhang2023hive, singh-etal-2023-image, El-Nouby_2019_ICCV,krojer2024learningactionreasoningcentricimage}.
Among recent work, InstructPix2Pix \citep{brooks2023instructpix2pixlearningfollowimage} leverages a training free editing model, Prompt-to-Prompt \citep{hertz2022prompt} along with
GPT3 \citep{NEURIPS2020_1457c0d6} to produce paired image data with editing instructions. MagicBrush \citep{zhang2024magicbrushmanuallyannotateddataset} is a manually annotated real world image editing dataset that improves 
InstructPix2Pix. \citep{krojer2024learningactionreasoningcentricimage} release Aurora, for action and reasoning centric image editing. In our work we explore the use of general-purpose image editors for aligning images generated via T2I models to their complex text prompts. We also develop a compositional image editor leading to further improved performance.



%% file: sec/3_method_1.tex
\section{Method}
In this section, we outline our \textit{Generate-Plan-Edit} (\algo) framework. Let us introduce some notation. Let $T$ be a textual instruction. In the task of T2I synthesis, given an instruction $T$, our goal is to be able to generate an image $I_o$ which satisfies various requirements expressed via the instruction $T$. While most existing techniques take the approach of directly generating $I_o$ via $T$, they often result in various kinds of inaccuracies, due to the complexity of the instruction. We are motivated by the observation that the task of T2I synthesis can be broken down into simpler steps of first generation, followed by identification of errors, and a sequence of corrective edits, each of which is simple and object specific in nature. Accordingly, we propose the following generation pipeline.

We first generate an initial image $I_g$ using a SOTA generative model, $\mathcal{G}$. This $I_g$ may have mistakes, or inaccuracies with respect to the intent expressed in $T$. We then make use of an existing MLLM ($\mathcal{P}$) to identify the mistakes in $I_g$, as a difference between image and textual descriptions in $T$ and $I_g$, respectively, for each object of interest, and corrective steps suggested, in the form of an edit plan expressed as $(T_{e_1},T_{e_2},\cdots, T_{e_n})$, where each $T_{e_k}$ is an edit instruction fixing some aspect of the $I_g$ so that it can be aligned with the original prompt $T$. Note that the number of edits is instance specific, and is a part of the output produced by $\mathcal{P}$. Finally, we make use of an editing model $\mathcal{E}$ which inputs the current image $I_{e_k}$ along with an edit instruction, $T_{e_{k+1}}$ and produces next image in the sequence $I_{e_{k+1}} = \mathcal{E}(I_{e_k},T_{e_{k+1}})$ with $k\in \{1,2,\cdots n-1\}$. Note that $I_g=I_{e_0}$ and $I_{e_n}=I_o$. This algorithm comprises of 3 broad steps: (a) {\em Generate}: Generate the image for textual instruction $\mathcal{T}$ 
(b) {\em Plan}: Identify mistakes and propose a correction plan (c) {\em Edit}: Perform the sequence of edits based on the corrective plan. We now describe each step in detail.

\begin{figure*}
  \centering
  \includegraphics[width=1.0\linewidth]{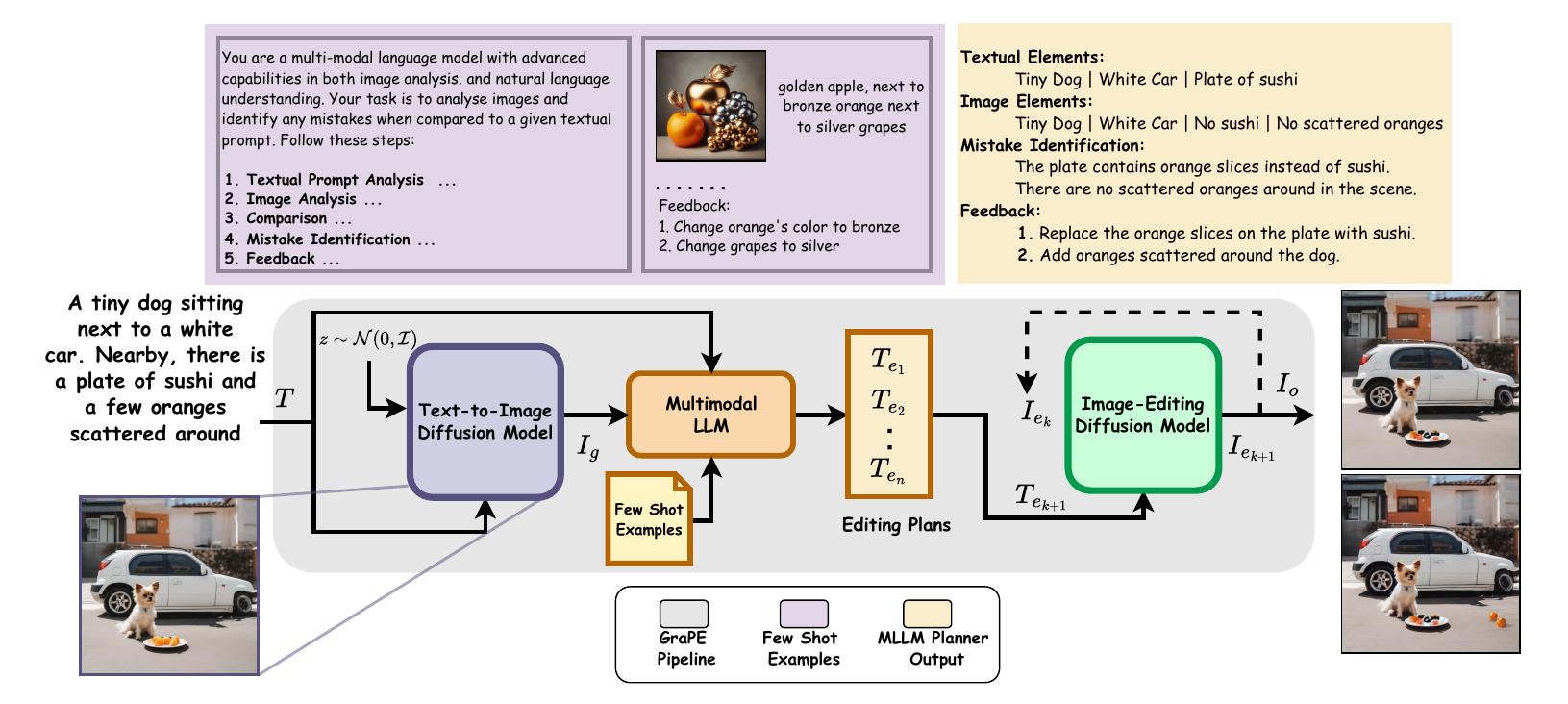}
  \caption{Proposed \algo\ framework, a given text prompt is used to generate an initial image from T2I model, $I_g$ which is then fed into a MLLM based planner along with the text prompt which identifies the objects that are misaligned in the image and outputs a set of edit plans guided by few-shot prompting. The plans are executed as a series of edits over the initial image to produce the final image}
            
  \label{fig:pipeline}
\end{figure*}

\subsection{Image Generation}
For our framework's initial step, we create a base image $I_g$, which serves as a foundational input for the subsequent processes involved in planning and editing. This image $I_{g}$ and it's corresponding text-prompt $T$ are considered as primary inputs for the next steps, which allows our framework to operate in a plug-and-play manner, regardless of the T2I model used in generation.


\subsection{Multi-Modal Planner}
The planner is a key component of \algo, with its role being executed by a Multi-Modal Large Language Model (MLLM) that assesses the initial image, $I_{g}$, for potential object misalignment relative to the textual description provided in the prompt , $T$. This discrepancy analysis acts similarly to the Chain-of-Thought prompting technique~\cite{wei2023chainofthoughtpromptingelicitsreasoning}, designed to enhance reasoning by encouraging the planner to break down the assessment process into object level steps.


\paragraph{Prompting Style:} The prompting style used with the MLLM play a crucial role in its ability to generate clear and concise plans across images from diverse domains. Unlike existing approaches \citep{lian2024llmgroundeddiffusionenhancingprompt, yang2024mastering}, which typically rely on LLMs to create abstract or structured representations--such as bounding boxes, layouts or blobs--either in a few-shot manner or through explicit training on annotated data \cite{zhang2024tierevolutionizingtextbasedimage}, our approach takes a streamlined alternative that bypasses the complexity of these strategies. 

Our method focuses on leveraging the inherent strengths of MLLMs, focusing on tasks where they already excel, such as image captioning and language comprehension, which reduces the need for intricate representations enabling our framework to maintain simplicity and generalization across domains. By combining these strengths with carefully designed few-shot examples, we guide the planner to produce object-centric, structured outputs without necessitating additional, specialized training.

The planner’s output is organized into four key steps, explained below using the example in Fig~\ref{fig:pipeline}.
\begin{itemize}
    \item \textbf{Analyzing Textual Elements:} The goal here is to extract high level object-attribute pairs from the text-prompt, $T$ focusing on relationships among the described objects. As shown in Fig.~\ref{fig:pipeline}, the key exacted entities are \textit{Tiny Dog}, \textit{White Car} \textit{and plate of sushi}.
    
    \item \textbf{Analyzing Image Elements:} Next, the planner generates a detailed object-level analysis of $I_g$, which is akin to creating a comprehensive image caption limited to object-level information. As seen in our example, MLLM effectively finds the objects present in the image in context of the text-prompt, $T$ i.e \textit{Tiny Dog} and \textit{White Car} and non-existence of \textit{sushi}, \textit{scattered oranges}.
    
    \item \textbf{Error Identification:} The extracted entities from both modalities extracted in the prior steps are compared and a descriptive summary of the identified errors is generated, this step not only grounds the mistakes but also provides an interpretable way into planner's reasoning.

    
    \item \textbf{Feedback Generation:} Finally, leveraging its extensive knowledge, the MLLM generates actionable feedback in the form of editing instructions to re-align image elements with the textual description. This step ensures that output aligns with the prompt's intent. As in Figure \ref{fig:pipeline}, the MLLM plans to first replaces the objects (orange slices) on the plate with sushi and adds oranges around the dog in two distinct steps.
 
\end{itemize}

This structured approach not only enhances the interpretability and accuracy of the planner’s outputs but also provides a user-friendly, transparent process for refining image alignments in \algo. The plans are extracted from MLLM output using regex matching of Feedback header.

\begin{table*}[h]
\centering
\scalebox{0.94}{
\begin{NiceTabular}{M{0.15\linewidth} | M{0.15\linewidth} M{0.15\linewidth} M{0.15\linewidth}}
\CodeBefore
    \Body
\toprule[1.5pt]
\textbf{Text Prompt} & \textbf{Generated Image} & \textbf{Edit Step 1} & \textbf{Edit Step 2} \\ \midrule


 \footnotesize{In the image, a corgi has a tiny apple positioned on top of its head. A cactus and a fork are placed nearby. Additionally, there is a duck with metallic texture in the scene.}
 & \includegraphics[width=25mm,height=25mm]{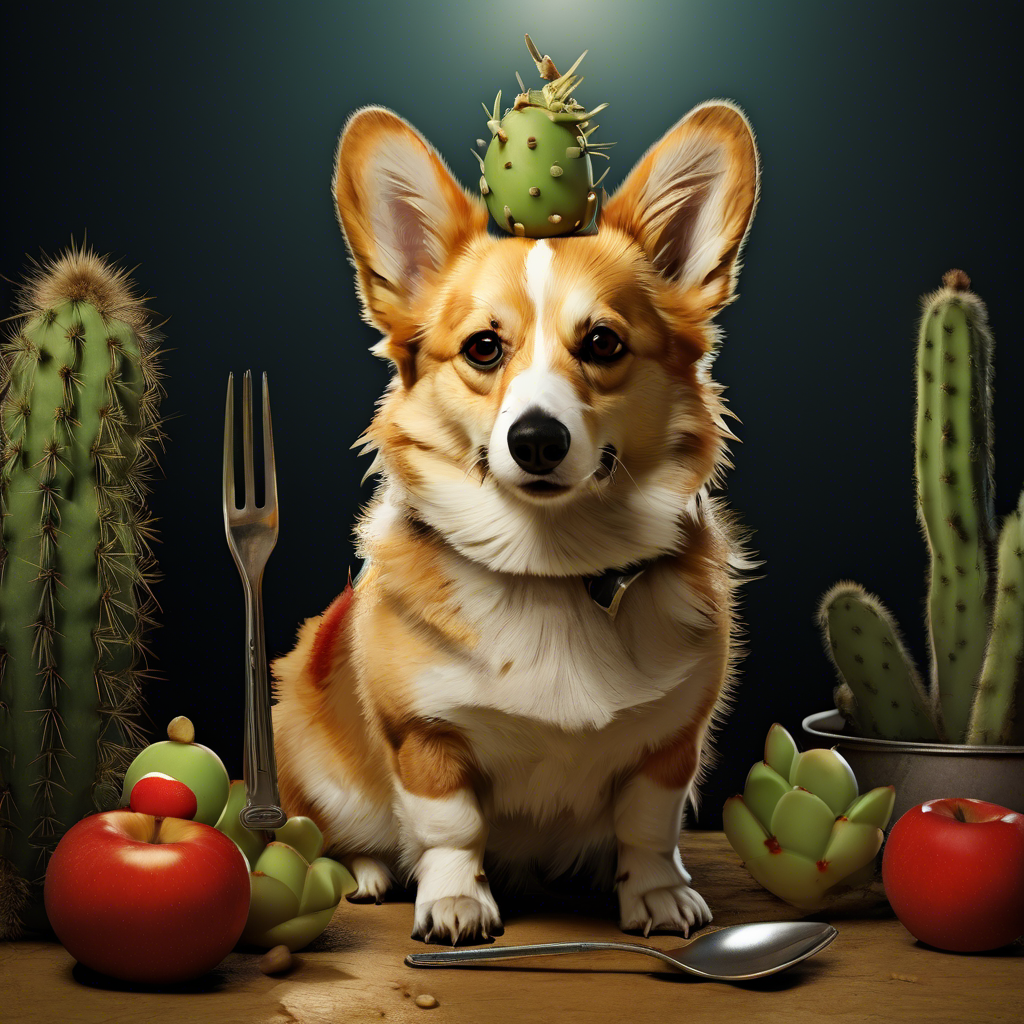} 
 & \includegraphics[width=25mm,height=25mm]{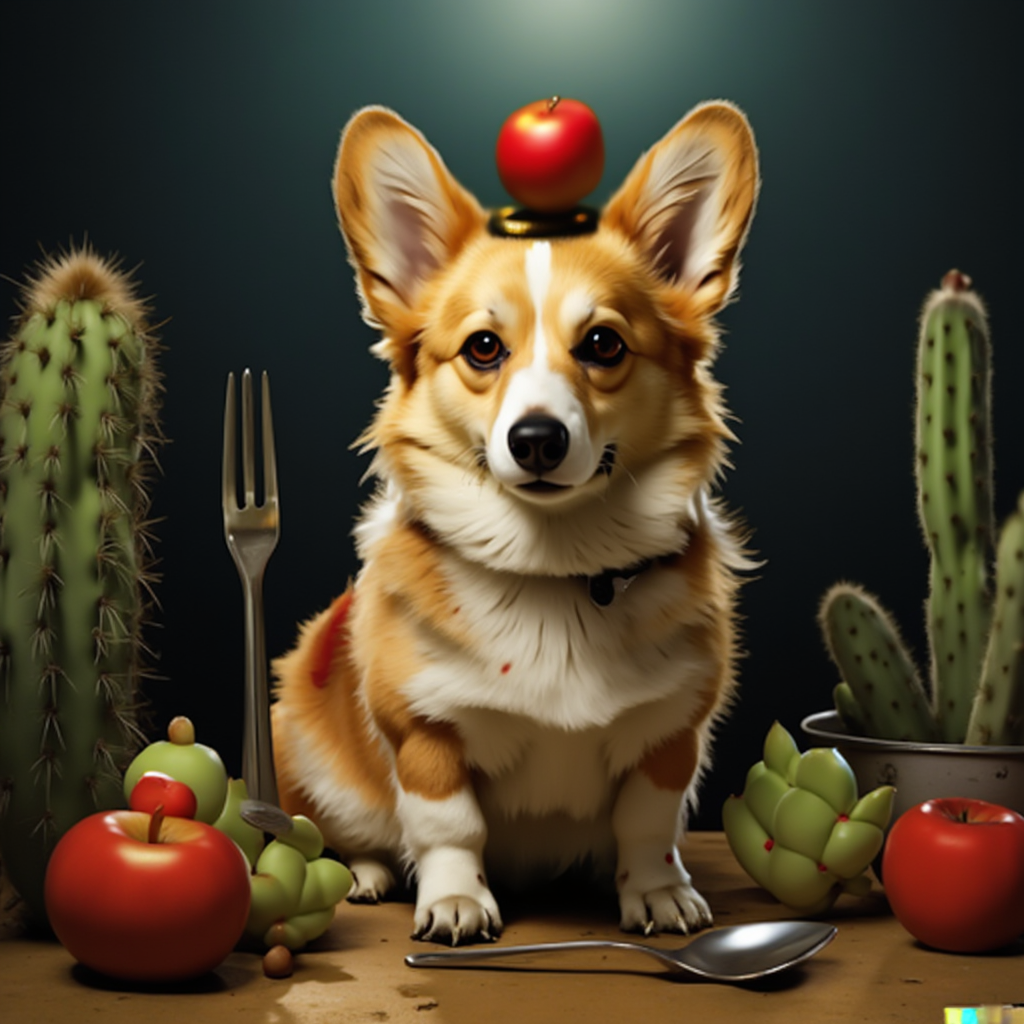} 
 & \includegraphics[width=25mm,height=25mm]{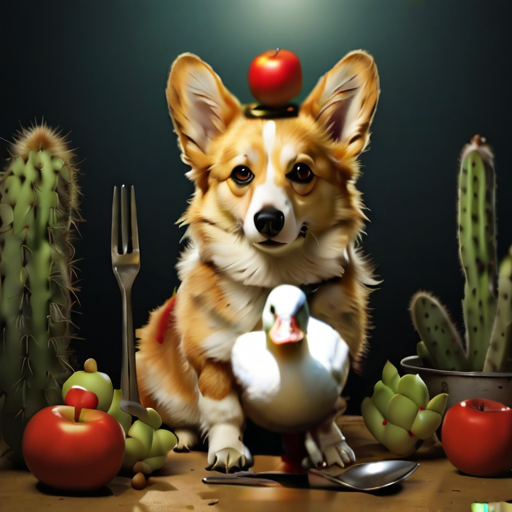} 
 \\
 &
 & \footnotesize{Replace the cactus on the corgi's head with a tiny apple}
 & \footnotesize{Add a duck with metallic texture to the scene} 
 \\ 
\midrule

 \footnotesize{A green bench, a red car, a blue bowl, and a pink apple. }
 & \includegraphics[width=25mm,height=25mm]{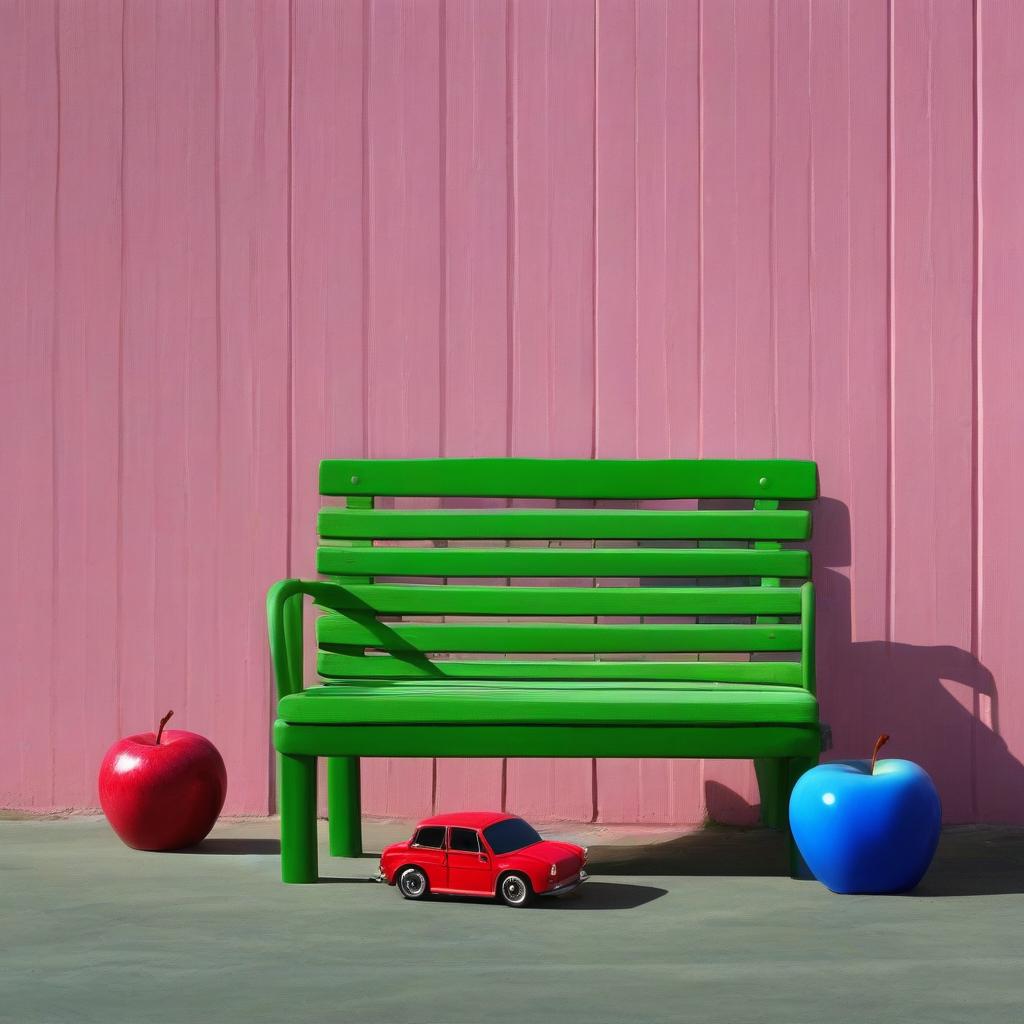} 
 & \includegraphics[width=25mm,height=25mm]{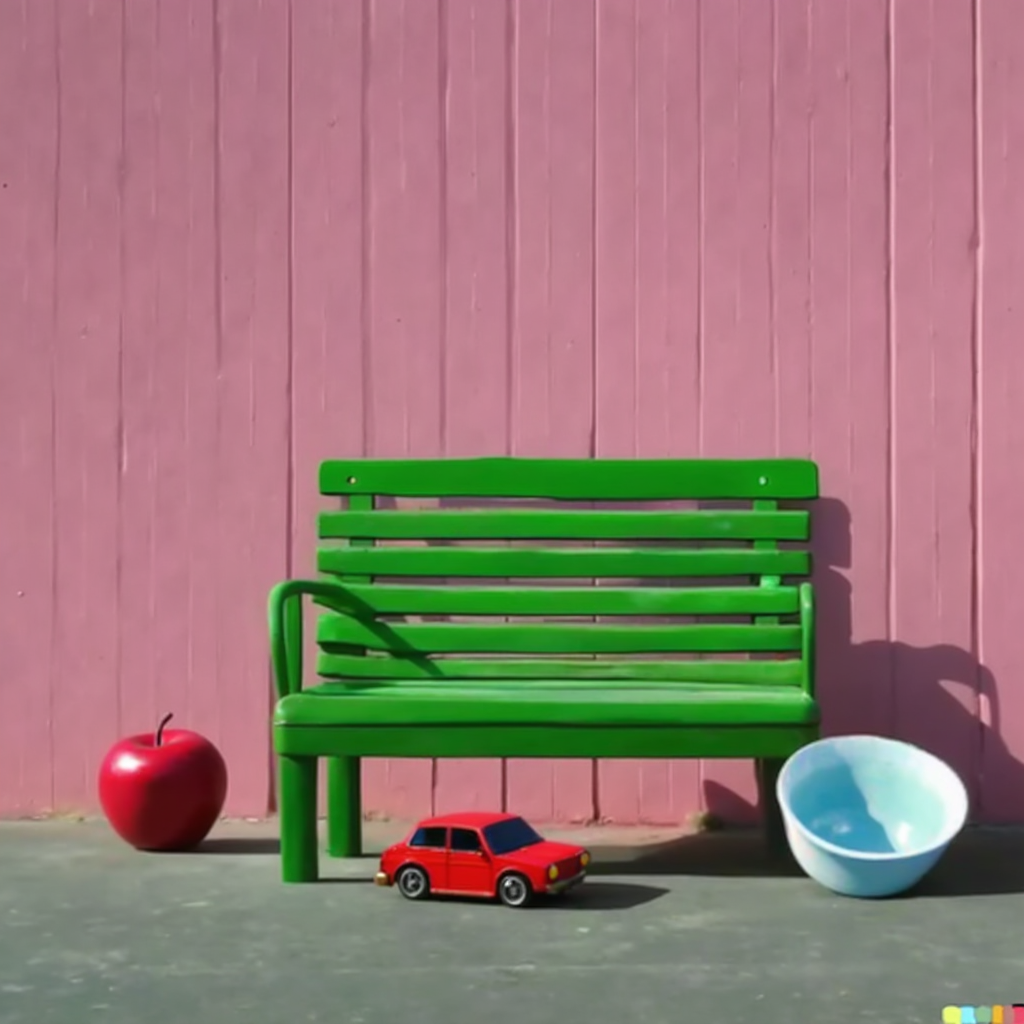} 
 & \includegraphics[width=25mm,height=25mm]{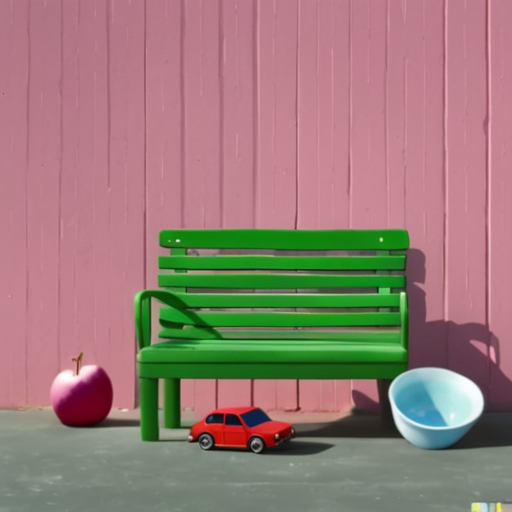} 
 \\
 &
 & \footnotesize{Change the blue apple to a blue bowl}
 & \footnotesize{Change the red apple to a pink apple}
 \\ 
\midrule


\footnotesize{A boy wearing a black shirt and khaki pants is laying down, and is surrounded by stuffed animals.}
 & \includegraphics[width=25mm,height=25mm]{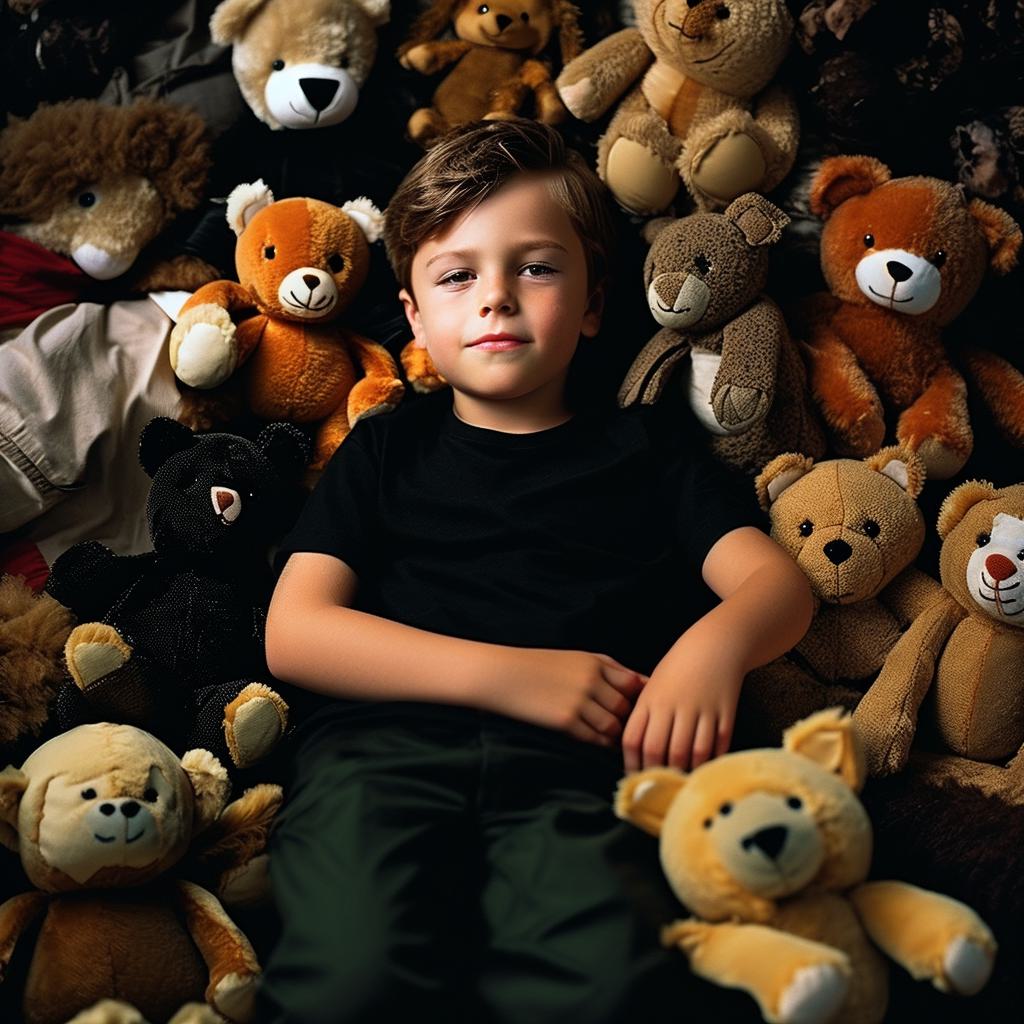} 
 & \includegraphics[width=25mm,height=25mm]{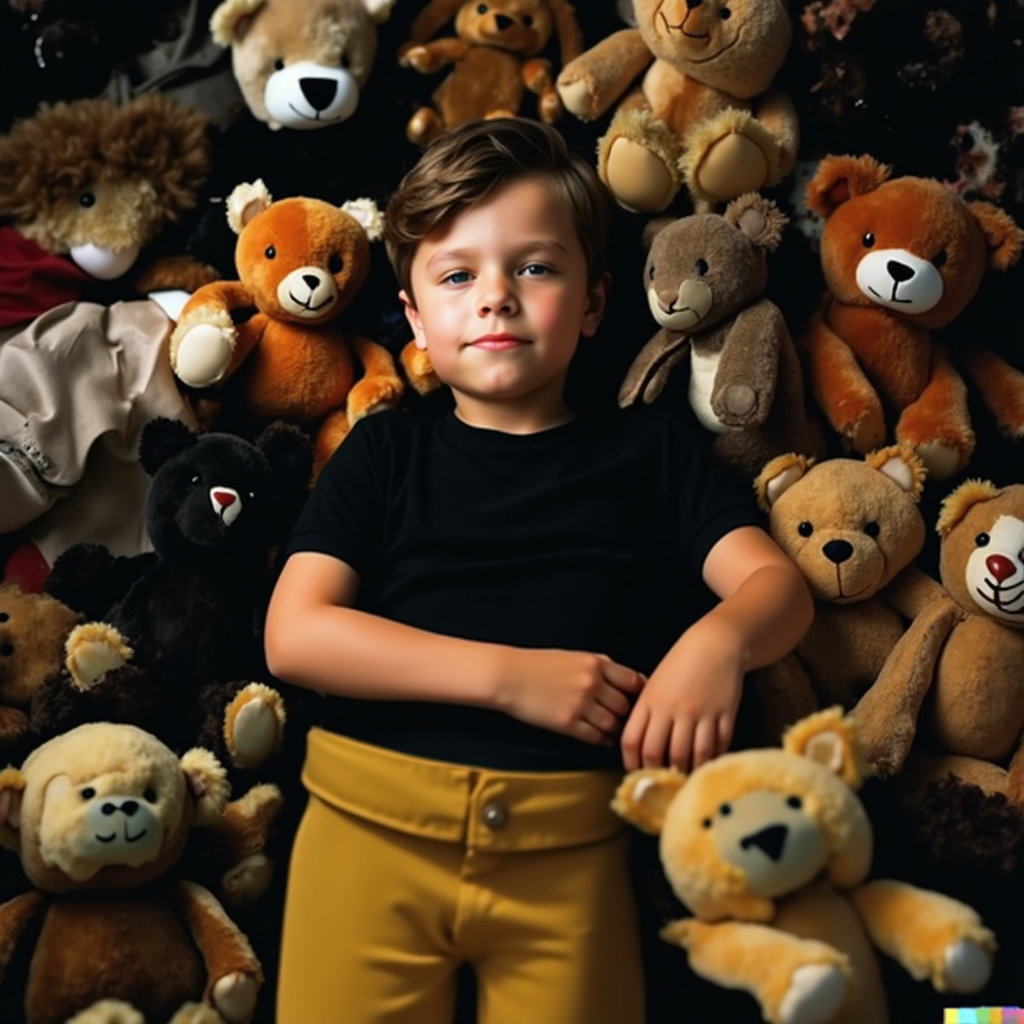} 
 &
 \\
 &
 & \footnotesize{Change the pants to khaki color}
 &
 \\ 



\bottomrule[1.5pt]
\end{NiceTabular}
}
\caption{Results produced by \algo. on images generated by SDXL, SD3.5-L and PixArt, using PixEdit as image-editor.} 
\vspace{-2mm}
\label{tab:customization}
\end{table*}

\subsection{Text Guided Image Editing}

In our framework, the editing model plays a crucial yet straightforward role: it iteratively refines the generated image ($I_g$) based on the plans extracted from the structured output produced by the MLLM in the planning phase. This design ensures flexibility, allowing for the use of any pre-trained model as a plug-and-play component.

\paragraph{Compositional Image Editing} T2I models based on CLIP~\cite{radford2021learningtransferablevisualmodels}, and by extension, editing models built on it are limited by CLIP’s compositional reasoning capabilities including understanding word order in text and object relations in images \citep{yuksekgonul2023visionlanguagemodelsbehavelike, kamath2023textencodersbottleneckcompositionality,Thrush2022WinogroundPV}.
We hypothesize and show for the first time that enhancing word order understanding in the text space and training the model on high-quality object-centric and reasoning-oriented data is key to induce compositionality in image editors.
For this, we introduce PixEdit, a text-guided image editing model based on PixArt-Sigma \cite{chen2024pixartsigmaweaktostrongtrainingdiffusion} and trained on the reasoning-centric dataset used by AURORA \citep{krojer2024learningactionreasoningcentricimage}. With the T5-XXL \cite{raffel2023exploringlimitstransferlearning} language model as its text encoder, PixEdit demonstrates enhanced performance on compositional and spatially complex edits, benefiting from both robust language comprehension and reasoning-centric training data. Providing gains over all existing image-editing models for the generation task in our \algo\ framework. Note that PixEdit is developed to be a \textit{general-purpose} image editor, with enhanced compositional and object centric editing capabilities, and is not specific to the generation task we tackle. Supp. Tables~\ref{tab:performance on AURORA-BENCH} and~\ref{tab:MB-TEST set performance} provide comparison of PixEdit's editing capability on two existing image editing benchmarks with other baselines. For detailed discussion about implementation and analysis of improvement in PixEdit refer Supp. Section \ref{supp:pixedit_implementation_details}.

\subsection{Implementation Details}
We utilize a frontier multi-modal model, GPT-4o as a multi-modal planner for its exceptional image-understanding and instruction following capabilities, both of which are essential for generating high-quality plans. While \algo\ works with any strong image editing model, for best results we use our developed PixEdit and the recently proposed AURORA model \cite{krojer2024learningactionreasoningcentricimage} as our editors due to their ability to perform object-centric and reasoning-oriented edits on real images, supported by training on an editing dataset derived from videos and simulators. See Supplementary Section \ref{supp:implementation_details} for more details on implementation.

%% file: sec/4_experiments.tex
\begin{figure*}
  \centering
  \includegraphics[width=1.0\linewidth]{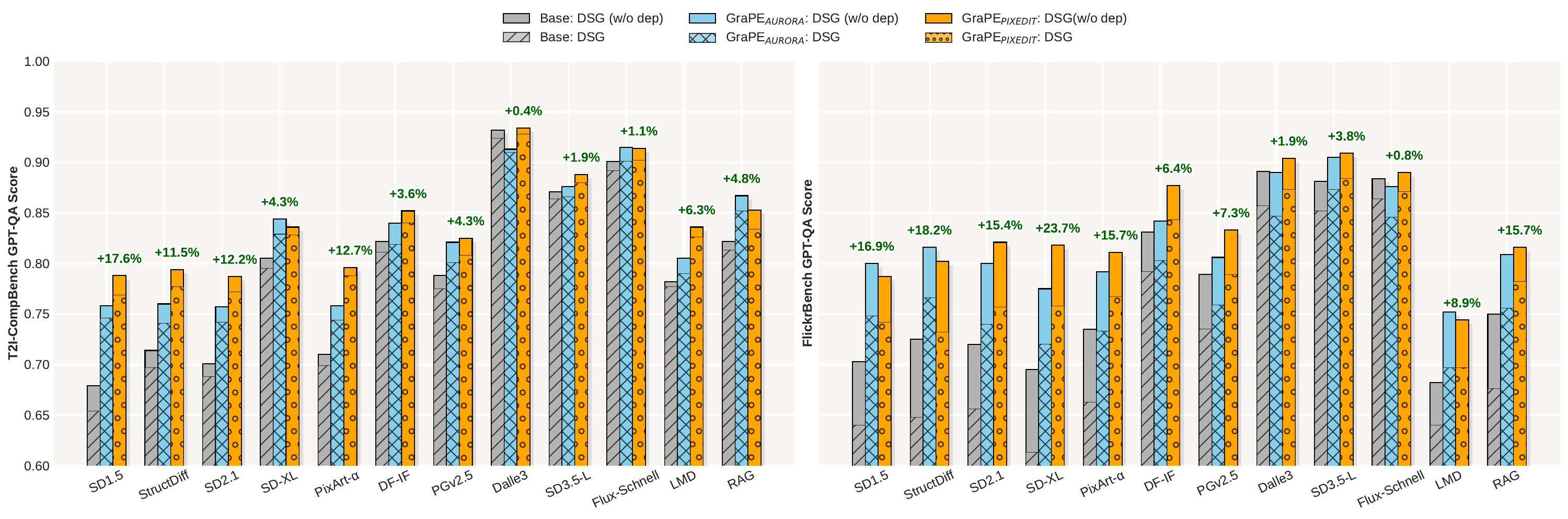}
    \caption{Experimental results showcasing the maximum gain in DSG score by \algo\ with both AURORA and PixEdit as editing models. The figure presents both DSG and DSG (w/o dependency) scores. The percentage gain is measured over DSG scores. The absolute values are presented in Table~\ref{table:t2i_compbench_bench_supp} and~\ref{tab:flickr_bench_detailed}}.
  \label{fig:flickr_bench_combined}
\end{figure*}

\begin{table*}[h!]
    \setlength{\tabcolsep}{6pt}
    \renewcommand{\arraystretch}{1.2}
    \centering
    \resizebox{\textwidth}{!}{
    \begin{tabular}{l |r r|r r|r r|r r}
    \toprule
    \multirow{2}{*}{Method} & \multicolumn{2}{c |}{Concept K=1} & \multicolumn{2}{c|}{Concept K=3} & \multicolumn{2}{c|}{Concept K=5} & \multicolumn{2}{c}{Concept K=7}\\ 
    \cline{2-9}
    & \multicolumn{1}{c}{ Base } & \multicolumn{1}{c|}{+\algo\textsubscript{PixEdit}} & \multicolumn{1}{c}{Base} & \multicolumn{1}{c|}{+\algo\textsubscript{PixEdit}} & \multicolumn{1}{c}{Base} & \multicolumn{1}{c|}{+\algo\textsubscript{PixEdit}} & \multicolumn{1}{c}{Base} & \multicolumn{1}{c}{+\algo\textsubscript{PixEdit}}\\ 
                                 
    \midrule                                                                        
    Stable-Diffusion v1.5 \citep{Rombach_2022_CVPR}                                      & $0.808$ \color{gray}$_{ \pm 0.009}$ 
                                                                                        & $\textbf{0.892}$ \color{gray}$_{\pm 0.002}$
                                                                                        & $0.606$ \color{gray}$_{ \pm 0.018}$ 
                                                                                        & $\textbf{0.752}$ \color{gray}$_{ \pm 0.002}$ 
                                                                                        & $0.497$ \color{gray}$_{ \pm 0.010}$ 
                                                                                        & $\textbf{0.689}$ \color{gray}$_{ \pm 0.002}$ 
                                                                                        & $0.450$ \color{gray}$_{ \pm 0.005}$ 
                                                                                        & $\textbf{0.610}$ \color{gray}$_{ \pm 0.005}$   \\
                                
    Structure Diffusion \citep{feng2023trainingfreestructureddiffusionguidance}          & $0.823$ \color{gray}$_{ \pm 0.002}$ 
                                                                                        & $\textbf{0.885}$ \color{gray}$_{\pm 0.004}$
                                                                                        & $0.606$ \color{gray}$_{ \pm 0.002}$
                                                                                        & $\textbf{0.723}$ \color{gray}$_{ \pm 0.005}$ 
                                                                                        & $0.542$ \color{gray}$_{ \pm 0.014}$ 
                                                                                        & $\textbf{0.703}$ \color{gray}$_{ \pm 0.006}$ 
                                                                                        & $0.447$ \color{gray}$_{ \pm 0.001}$ 
                                                                                        & $\textbf{0.571}$ \color{gray}$_{ \pm 0.002}$  \\
                                
    Stable Diffusion v2.1 \citep{Rombach_2022_CVPR}                                      & $0.833$ \color{gray}$_{ \pm 0.002}$ 
                                                                                        & $\textbf{0.868}$ \color{gray}$_{\pm 0.005}$
                                                                                        & $0.639$ \color{gray}$_{ \pm 0.014}$ 
                                                                                        & $\textbf{0.737}$ \color{gray}$_{ \pm 0.002}$ 
                                                                                        & $0.579$ \color{gray}$_{ \pm 0.012}$ 
                                                                                        & $\textbf{0.687}$ \color{gray}$_{ \pm 0.005}$ 
                                                                                        & $0.466$ \color{gray}$_{ \pm 0.002}$ 
                                                                                        & $\textbf{0.626}$ \color{gray}$_{ \pm 0.002}$ \\

    SD-XL \citep{podell2023sdxlimprovinglatentdiffusion}                                 & $0.848$ \color{gray}$_{ \pm 0.010}$ 
                                                                                        & $\textbf{0.877}$ \color{gray}$_{\pm 0.005}$
                                                                                        & $0.708$ \color{gray}$_{ \pm 0.018}$
                                                                                        & $\textbf{0.780}$ \color{gray}$_{ \pm 0.007}$ 
                                                                                        & $0.635$ \color{gray}$_{ \pm 0.014}$ 
                                                                                        & $\textbf{0.729}$ \color{gray}$_{ \pm 0.004}$ 
                                                                                        & $0.520$ \color{gray}$_{ \pm 0.003}$ 
                                                                                        & $\textbf{0.628}$ \color{gray}$_{ \pm 0.002}$  \\
    
    PixArt-$\alpha$ \citep{chen2023pixartalphafasttrainingdiffusion}                     & $0.813$ \color{gray}$_{ \pm 0.010}$ 
                                                                                        & $\textbf{0.872}$ \color{gray}$_{\pm 0.010}$
                                                                                        & $0.668$ \color{gray}$_{ \pm 0.011}$ 
                                                                                        & $\textbf{0.722}$ \color{gray}$_{ \pm 0.002}$ 
                                                                                        & $0.649$ \color{gray}$_{ \pm 0.011}$
                                                                                        & $\textbf{0.742}$ \color{gray}$_{ \pm 0.002}$ 
                                                                                        & $0.507$ \color{gray}$_{ \pm 0.001}$ 
                                                                                        & $\textbf{0.625}$ \color{gray}$_{ \pm 0.002}$  \\
    
    DeepFloyd IF \citep{DeepFloydIF}                                                     & $0.883$ \color{gray}$_{ \pm 0.009}$ 
                                                                                        & $\textbf{0.915}$ \color{gray}$_{\pm 0.000}$
                                                                                        & $0.680$ \color{gray}$_{ \pm 0.016}$ 
                                                                                        & $\textbf{0.765}$ \color{gray}$_{ \pm 0.007}$ 
                                                                                        & $0.663$ \color{gray}$_{ \pm 0.014}$
                                                                                        & $\textbf{0.745}$ \color{gray}$_{ \pm 0.002}$ 
                                                                                        & $0.583$ \color{gray}$_{ \pm 0.002}$ 
                                                                                        & $\textbf{0.662}$ \color{gray}$_{ \pm 0.006}$ \\
    
    PlaygroundV2.5 \citep{li2024playgroundv25insightsenhancing}                          & $0.908$ \color{gray}$_{ \pm 0.010}$ 
                                                                                        & $\textbf{0.955}$ \color{gray}$_{\pm 0.004}$
                                                                                        & $0.737$ \color{gray}$_{ \pm 0.023}$
                                                                                        & $\textbf{0.792}$ \color{gray}$_{ \pm 0.009}$ 
                                                                                        & $0.658$ \color{gray}$_{ \pm 0.015}$
                                                                                        & $\textbf{0.721}$ \color{gray}$_{ \pm 0.002}$ 
                                                                                        & $0.540$ \color{gray}$_{ \pm 0.003}$ 
                                                                                        & $\textbf{0.640}$ \color{gray}$_{ \pm 0.005}$ \\
    
    Dalle3 \citep{betker2023improving}                                                   & $0.947$ \color{gray}$_{ \pm 0.002}$ 
                                                                                        & $\textbf{0.953}$ \color{gray}$_{\pm 0.002}$
                                                                                        & $0.832$ \color{gray}$_{ \pm 0.012}$ 
                                                                                        & $\textbf{0.861}$ \color{gray}$_{ \pm 0.003}$ 
                                                                                        & $0.812$ \color{gray}$_{ \pm 0.014}$ 
                                                                                        & $\textbf{0.832}$ \color{gray}$_{ \pm 0.004}$ 
                                                                                        & $0.728$ \color{gray}$_{ \pm 0.006}$ 
                                                                                        & $\textbf{0.737}$ \color{gray}$_{ \pm 0.004}$  \\
    
    Stable Diffusion v3.5 Large \citep{sd3_5_dummy}                                                 & $0.927$ \color{gray}$_{ \pm 0.005}$ 
                                                                                        & $\textbf{0.948}$ \color{gray}$_{\pm 0.002}$
                                                                                        & $0.815$ \color{gray}$_{ \pm 0.002}$ 
                                                                                        & $\textbf{0.817}$ \color{gray}$_{ \pm 0.002}$ 
                                                                                        & $0.803$ \color{gray}$_{ \pm 0.003}$ 
                                                                                        & $\textbf{0.831}$ \color{gray}$_{ \pm 0.004}$ 
                                                                                        & $0.759$ \color{gray}$_{ \pm 0.004}$ 
                                                                                        & $\textbf{0.784}$ \color{gray}$_{ \pm 0.005}$ \\

    Flux-schnell \citep{flux}                                                 & $0.902$ \color{gray}$_{ \pm 0.002}$ 
                                                                                        & $\textbf{0.918}$ \color{gray}$_{\pm 0.002}$
                                                                                        & $0.820$ \color{gray}$_{ \pm 0.003}$ 
                                                                                        & $\textbf{0.864}$ \color{gray}$_{ \pm 0.001}$ 
                                                                                        & $0.786$ \color{gray}$_{ \pm 0.005}$ 
                                                                                        & $\textbf{0.804}$ \color{gray}$_{ \pm 0.008}$ 
                                                                                        & $0.775$ \color{gray}$_{ \pm 0.004}$ 
                                                                                        & $\textbf{0.779}$ \color{gray}$_{ \pm 0.004}$ \\

    LMD \citep{lian2024llmgroundeddiffusionenhancingprompt}                              & $0.855$ \color{gray}$_{ \pm 0.004}$ 
                                                                                        & $\textbf{0.873}$ \color{gray}$_{\pm 0.002}$
                                                                                        & $0.711$ \color{gray}$_{ \pm 0.008}$ 
                                                                                        & $\textbf{0.773}$ \color{gray}$_{ \pm 0.006}$ 
                                                                                        & $0.643$ \color{gray}$_{ \pm 0.011}$
                                                                                        & $\textbf{0.725}$ \color{gray}$_{ \pm 0.009}$ 
                                                                                        & $0.591$ \color{gray}$_{ \pm 0.002}$ 
                                                                                        & $\textbf{0.668}$ \color{gray}$_{ \pm 0.009}$  \\

    RAG \citep{chen2024regionaware}                                                            & $0.815$ \color{gray}$_{ \pm 0.009}$ 
                                                                            & $\textbf{0.866}$ \color{gray}$_{\pm 0.003}$
                                                                            & $0.668$ \color{gray}$_{ \pm 0.007}$ 
                                                                            & $\textbf{0.718}$ \color{gray}$_{ \pm 0.005}$ 
                                                                            & $0.665$ \color{gray}$_{ \pm 0.002}$ 
                                                                            & $\textbf{0.721}$ \color{gray}$_{ \pm 0.004}$ 
                                                                            & $0.520$ \color{gray}$_{ \pm 0.003}$ 
                                                                            & $\textbf{0.628}$ \color{gray}$_{ \pm 0.001}$ \\

    RPG \citep{yang2024mastering}                                  & $0.696$ \color{gray}$_{ \pm 0.015}$ 
                                                                            & $\textbf{0.845}$ \color{gray}$_{\pm 0.008}$
                                                                            & $0.694$ \color{gray}$_{ \pm 0.002}$ 
                                                                            & $\textbf{0.715}$ \color{gray}$_{ \pm 0.007}$ 
                                                                            & $0.583$ \color{gray}$_{ \pm 0.002}$ 
                                                                            & $\textbf{0.688}$ \color{gray}$_{ \pm 0.005}$ 
                                                                            & $0.388$ \color{gray}$_{ \pm 0.007}$ 
                                                                            & $\textbf{0.467}$ \color{gray}$_{ \pm 0.005}$ \\

    \bottomrule
    \end{tabular}
    }
    \caption{Results on Concept-mix benchmark ~\algo\textsubscript{PixEdit}. For complementary table with AURORA as editing model refer Table~\ref{tab:conceptmix_detailed_aurora}.}
    \label{table:concept_mix_pixedit}
    \end{table*}

\section{Experiments}
\subsection{Experimental Settings}
\paragraph{Models:} To evaluate the effectiveness of \algo, we conduct a comprehensive assessment across 13 state-of-the-art text-to-image (T2I) models of varying sizes and capabilities. These models include: Stable Diffusion v1.5 \citep{Rombach_2022_CVPR}, Structure Diffusion \citep{feng2023trainingfreestructureddiffusionguidance}, Stable Diffusion V2.1 \citep{Rombach_2022_CVPR}, SD-XL \citep{podell2023sdxlimprovinglatentdiffusion}, PixArt-$\alpha$ \citep{chen2023pixartalphafasttrainingdiffusion}, DeepFloyd IF \citep{DeepFloydIF}, PlaygroundV2.5 \citep{li2024playgroundv25insightsenhancing}, Dalle 3 \citep{betker2023improving}, Stable Diffusion v3.5 Large \citep{sd3_5_dummy}, Flux-schnell~\citep{flux} and three baselines which adopts MLLMs for complex T2I generation: LMD \citep{lian2024llmgroundeddiffusionenhancingprompt}, RPG~\cite{yang2024mastering} and RAG~\cite{chen2024regionaware}. We also compare \algo~with recently proposed SLD~\cite{wu2023self} in Suppl. Section~\ref{sec:SLD}, which is a post-hoc self correction method using MLLMs. This selection encompasses a diverse range of recent T2I models, allowing for a thorough evaluation of \algo 's robustness across different architectures and configurations.

\paragraph{Benchmarks:}


\begin{enumerate}
    \item \textbf{T2I Compbench} \citep{huang2023t2icompbenchcomprehensivebenchmarkopenworld}: We evaluate the compositional generation capabilities of \algo\ using this well-established benchmark, specifically designed to assess performance in this area. This includes compositional challenges across categories such as shape, color, texture, spatial, non-spatial, and complex. We utilize a subset of 100 prompts, sampled uniformly across all six categories—shape, color, texture, spatial, non-spatial, and complex—to ensure a comprehensive assessment of the model's performance across diverse compositional challenges.
    
    \item \textbf{ConceptMix} \citep{wu2024conceptmixcompositionalimagegeneration} : This benchmark evaluates models across varying levels of controllable compositionality. Using the official ConceptMix codebase, we generate 100 prompts for each $K \in {1, 3, 5, 7}$. Each prompt includes at least one object paired with $K$ additional visual concepts—such as color, number, shape, size, style, spatial arrangement, and texture—offering a diverse and rigorous assessment of the model's ability to handle increasingly complex visual compositions.
    \item \textbf{Flickr-Bench} : Flickr-30k \citep{plummer2016flickr30kentitiescollectingregiontophrase} is a widely used benchmark for evaluating the generation quality of T2I models using metrics such as FID \citep{heusel2018ganstrainedtimescaleupdate}. The human-generated prompts in this dataset offer a well-balanced mix of compositionality and realism, making it suitable for assessing our method's performance on general-purpose tasks beyond compositionality. To this end, we sample 100 prompts from the dataset's test split for evaluation.
\end{enumerate}

\paragraph{Evaluation Metrics:} 
Many of the SOTA automated evaluation metrics belong to the QA pair generation and Visual Question-Answering(VQA) framework. Prior works \citep{hu2023tifaaccurateinterpretabletexttoimage, yarom2023readimprovingtextimagealignment, cho2024davidsonianscenegraphimproving,gokhale2023benchmarkingspatialrelationshipstexttoimage, wiles2024revisitingtexttoimageevaluationgecko} have demonstrated it's effectiveness and correlation with human-predictions when judging semantic accuracies. The ability to generate object/attribute centric 'yes'/'no' questions over a compositional text prompt allows for fine-grained assessment of image elements using MLLMs. We therefore follow \citep{cho2024davidsonianscenegraphimproving} to generate grounded binary questions for T2I-Compbench and Flickr-Bench and utilize ConceptMix's existing set of questions that are generated simultaneously with the prompts and utilize GPT-4o as the choice VQA model following \cite{wu2024conceptmixcompositionalimagegeneration}.

\subsection{Results}
\noindent\textbf{\algosim\ Improves Compositional T2I Synthesis}
\algo\ can be used with any given back-bone in a plug-n-play manner, and substantially improves performance across the board for all of the 13 T2I models tested in this study, including SoTA models like SD3.5 \citep{sd3_5_dummy}. Results on T2I Compbench and Flickr-Bench are present in Figure \ref{fig:flickr_bench_combined}. Results on ConceptMix are present in Table \ref{table:concept_mix_pixedit}. On T2I Compbench we see improvements ranging from 17.6\% in the case of SD1.5 to nearly 2\% for the latest SD3.5 Large model. On ConceptMix as well, we see large performance gains across prompt complexities $K$=$\{1,3,5,7\}$. For prompts with higher complexity ($K$=$7$), we see gains ranging from 3.1\% to 35.5\% across models.

\noindent\textbf{\algosim\ is more effective with complex T2I prompts.}
As seen in Table \ref{table:concept_mix_pixedit}, the absolute performance difference between generation from base model vs our method increases as we go towards more complex prompts (from $K$=$1$ to $K$=$7$). For e.g., for SD XL, the gains compared to the base model increase from 2.9\% for $K$=$1$ to 10.8\% for $K$=$7$, and this observation remains constant across T2I models.

\noindent\textbf{\algosim\ exceeds performance on general T2I tasks}
Gains achieved by \algo\ are not limited to compositional or complex prompts as is the case in T2I Compbench and ConceptMix datasets, rather it also generalizes to more widely used general T2I benchmarks such as Flickr-Bench. Fig. \ref{fig:flickr_bench_combined} shows that \algo\ leads to large performance gains ranging from 2 to 23\% across different models.

\noindent\textbf{\algosim\ reduces the gap between models of varied T2I capabilities}
\algo\ leads to a larger absolute performance increase for relatively less capable base diffusion models. This closes the gap between performance of such models with more capable ones. For e.g., for ConceptMix dataset ($K$=$3$), the difference between SD-1.5 and SD-3.5 reduces from 20.9\% to 6.5\% when comparing the base generations vs generations from \algo\ respectively.

\begin{figure}[t]
  \centering
   \includegraphics[width=1.0\linewidth]{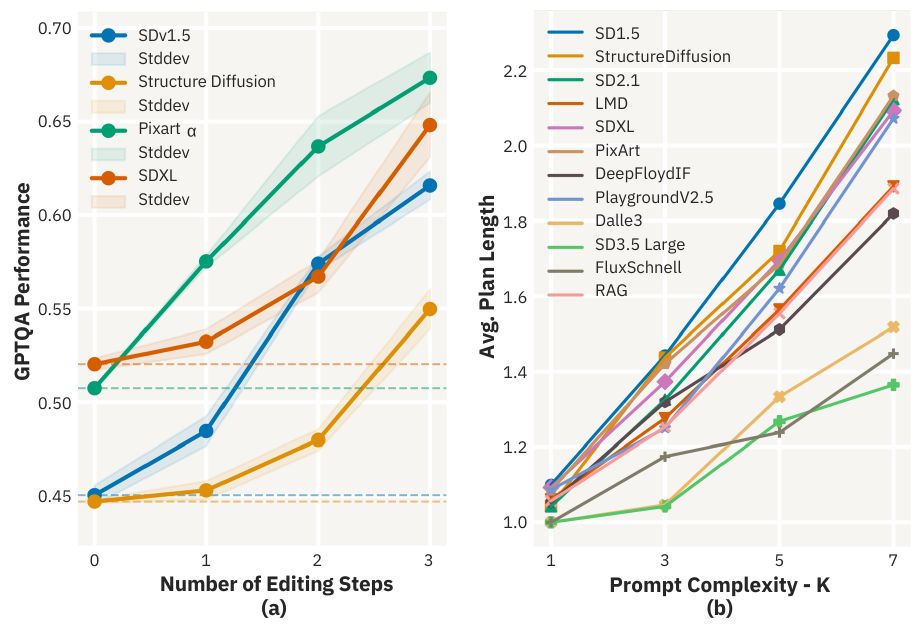}
   \caption{(a) Trend of GPT-QA score with increasing number of editing steps. (b) Average edit steps per plan for ConceptMix benchmark across the models}
   \label{fig:perf vs editing steps}
\end{figure}

\noindent\textbf{\algosim\ scales well with more compute}
Our approach can be viewed from a lens of using extra inference time compute (calls to the editing model) for creating images better aligned with complex prompts, achieving better performance on the T2I Synthesis task. \algo\ flexibly trades off performance with the amount of compute (or the number of edits at inference time). See Fig. \ref{fig:perf vs editing steps}(a) for results on four representative models. This also shows that each planning step of \algo\ is important on average for improving on the T2I synthesis task gradually. 


\subsection{Ablations}
\label{naive_planner_ablation}
\paragraph{Naive Planner}
This section highlights the importance of object and attribute centric decomposition of plans to generate effective editing instructions. We modify the few-shot structure of the MLLM planner to skip the decomposition into image and text elements and prompt it to identify errors and suggest editing plans given the text-prompt and generated image. We refer to the pipeline thus constructed as GraPE$_{naive}$. Table \ref{table:concept_mix_pixedit_grape_naive} presents the results of both \algo\ and GraPE$_{naive}$ on $K$=$7$ subset of ConceptMix benchmark for a subset of models. See Supp. Section \ref{supp:ablations} for further details.

\begin{table}[h!]
    \resizebox{\columnwidth}{!}{
    \centering
    \begin{tabular}{l |r r}
    \toprule
    \multirow{2}{*}{Method} & \multicolumn{2}{c}{Concept K=7}\\ 
    \cline{2-3}
    & \multicolumn{1}{c}{GraPE$_{naive}$} & \multicolumn{1}{c}{GraPE}\\ 
                                 
    \midrule                                                                        
                                
                                
    Stable Diffusion v2.1 \citep{Rombach_2022_CVPR}                                
                                                                                       & $0.618$ 
                                                                                        & $\textbf{0.626}$ \\
    
    LMD \citep{lian2024llmgroundeddiffusionenhancingprompt}                             
                                                                                        & $0.622$ 
                                                                                        & $\textbf{0.668}$ \\
    
    SD-XL \citep{podell2023sdxlimprovinglatentdiffusion}                               
                                                                                        & $0.598$  
                                                                                        & $\textbf{0.628}$ \\
    
    
    
    PlaygroundV2.5 \citep{li2024playgroundv25insightsenhancing}                          
                                                                                        & $0.622$ 
                                                                                        & $\textbf{0.640}$ \\
    
    
    Stable Diffusion v3.5 Large \citep{sd3_5_dummy}                                 
                                                                                       & $0.724$ 
                                                                                        & $\textbf{0.784}$ \\ 
    
    \bottomrule
    \end{tabular}
    }
    \caption{Results on Concept-mix benchmark ($K$=$7$ subset). Comparing both GraPE$_{naive}$ and \algo~(Using PixEdit)}
    \label{table:concept_mix_pixedit_grape_naive}
    \end{table}

\subsection{Analysis and Insights}
We conduct a detailed examination of each component in \algo, focusing specifically on identifying and analyzing failure cases in both the MLLM planner and the editing model. For this purpose, we curate a dataset of 120 image-prompt pairs, uniformly sampled across all 13 models and 6 benchmarks. This dataset is carefully evaluated by six human annotators, compensated well above the national average hourly wage.

\paragraph{MLLM planner Accuracy}


\begin{table}[h!]
    \centering
    \small 
    \renewcommand{\arraystretch}{1.2} 
    \setlength{\tabcolsep}{4pt} 
    \begin{tabular}{@{}lccccccc@{}}
        \toprule
        & \multicolumn{2}{c}{Q1} & \multicolumn{3}{c}{Q2} & \multicolumn{2}{c}{Q3} \\
        \cmidrule(lr){2-3} \cmidrule(lr){4-6} \cmidrule(lr){7-8}
        Type & Yes & No & No & Partial & Full & Yes & No \\
        \midrule
        GPT-4o & 25.4 & 74.5 & 8.7 & 35.7 & 55.5
 & 89.4 & 10.5 \\
        Qwen-VL-72B & 25.4 & 74.5 & 26.9 & 30.0 & 43.0 & 85.9 & 14.0\\
        \bottomrule
    \end{tabular}
    \caption{Results from human study on MLLM planners. We present results with GPT-4o based planner used in \algo\ and an open-source alternative Qwen-VL-72B~\cite{bai2023qwenvlversatilevisionlanguagemodel}}
    \label{table:human_eval}
\end{table}

For each image-prompt pair, along with the \textit{plan} generated by \algo, evaluators were asked to answer three multiple-choice questions:
\begin{itemize}
    \item Q1: Does the generated image accurately and completely match the content described in the text prompt?
    \item Q2: To what extent does executing the proposed plan improve the image’s alignment with the text prompt?
    \item Q3: Are the instructions in the plan indivisible, meaning each instruction is a fundamental action that cannot be further subdivided?
\end{itemize}

Questions Q1 and Q3 require binary answers, with 'yes' (1) or 'no' (0) responses. Q2 offers three response options: 'no improvement', 'partial improvement' and 'full alignment'. This structured evaluation allows us to assess the accuracy and granularity of the MLLM planner in relation to image-prompt alignment.


Our analysis reveals that Qwen-VL-72B as planner has around 18\% more cases of 'no improvement' compared to GPT-4o, largely due to generating a considerable number of empty plans. This suggests that while Qwen-VL-72B may trail GPT-4o in visual capabilities it often fails to identify discrepancies, but tends to produce a solid plan when it does recognize them, as evident from Table \ref{table:human_eval}. We also carry out the annotation process on a smaller model, Qwen-VL-7B which in contrast to the above planners performed poorly, generating approximately 100 empty plans out of 120 images, indicating significant limitations in both identifying discrepancies and generating actionable plans.

\begin{figure}
  \centering
  \includegraphics[width=1.0\linewidth]{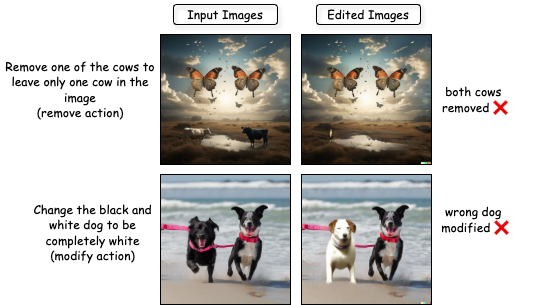}
  \caption{Results illustrating failure cases of Image-Editing model}
  \label{fig:error_analysis}
\end{figure}

\paragraph{Error Analysis of Editing Model}
We qualitatively take a look at the failure cases of the editing model (PixEdit) specifically utilizing the above dataset to gather a subset of plans which are deemed to fully align the image with the text prompt by the majority of evaluators (based on Q2 earlier). This subset consists of about 46 plans in which any errors in the final image solely result from the shortcomings of the image-editing model. We specifically look at images in which correct, partial or incorrect edits were performed by the model. Out of these 14 were completely correct, 17 partially correct, and 15 were incorrect. Within the correct ones, the distribution of correct add, remove and modify was 12, 0 and 2, respectively. Within the partially correct ones, the distribution of correct add, remove and modify was 11, 2 and 4, respectively. Figure~\ref{fig:error_analysis} provides some representative examples. Fixing these issues, possibly through an RL based mechanism for incorporating feedback is a direction for future work. This also points to the fact that the primary reason for any errors originates from the editing model, rather than because of the planner. 



\paragraph{Multi-Modal Planner as Evaluation Metric}
Can planning help evaluating T2I generative models? For this, we look at the analysis and mistake identification capabilities of the planner. Specifically, we post-process the planner-generated output filtering out only the textual-image element comparison and pass this back to the planner which is provided a zero-shot system prompt. This prompt directs the planner to generate a score on a 1-100 scale identifying the alignment based on the objects, their relationships and attributes. When compared  with LLMScore~\citep{lu2023llmscore}, We observed a moderate correlation of 0.423 and 0.404 when tested using Spearman and Pearson's test respectively. This shows the future potential of our MLLM planner to be used in conjunction with SoTA evaluation metrics to further improve them.

\paragraph{Number of planning steps}
In Fig.~\ref{fig:perf vs editing steps}(b), we see a gradual increase in the average number of plans as we increase the prompt complexity from $K$=$1$ to $K$=$7$ using the ConceptMix benchmark. This shows our planner is able to recognize the complexity of the prompt and generate a longer plan in response. We also see that on average, more capable T2I models like SD3.5 require fewer editing steps compared to weaker models such as SD1.5. 

For inference cost analysis and runtime comparison refer to Suppl. Section~\ref{supp:more_results}. Additional qualitative examples of our approach are shown in Suppl. Section \ref{supp:qualitative}.


%% file: sec/5_conclusion.tex
\section{Discussion and Limitations}
\textbf{Summary:} In this paper, we have looked at the task of fine-grained image generation for compositional prompts. Unlike most exiting methods, which try to achieve this task in one-go, we take a different route, and decompose the problem in 3-steps: (a) Generate an initial image based on the prompts (b) Identify the mistakes and plan out corrective steps by anaylzing the difference between textual and visual elements for each object (c) Execute the corrective steps via an editing model in a sequential manner. The ability of our model to break the corrective steps into multiple simple edits is key to the success of our approach. Extensive experimentation on multiple datasets shows that our method can significantly improve the generation quality over 10 different models, including SoTA approaches. 

\noindent\textbf{Limitations and future work:} Limitations of our approach include its inability to handle certain kinds of complex prompts, and its dependence on a pre-trained MLLM in the planning step. Future work includes working with more complex prompts, integrating our approach with models which explicitly change the architecture to better align with text, and incorporating corrective feedback using an RL based framework and extending to image editing tasks.

%% file: sec/X_suppl.tex
\clearpage
\setcounter{page}{1}
\maketitlesupplementary

\section{Algorithm:}
\label{supp:algorithm}
We present the algorithm showcasing the sequence of steps in \algo\ below.
\begin{algorithm}
\caption{Generate-Plan-Edit : \algosim }\label{alg:GraPE}
\begin{algorithmic}
\Require Text Prompt: $T$, T2I Model: $G$, MLLM: $\mathcal{P}$, Editing Model: $\mathcal{E}$ and Few-Shot Examples: $\big[E_1$ $\cdots$ $E_{p}\big]$\\
\textit{\textcolor{gray}{\#Generate}}
\State $I_g \gets \mathcal{G}(T)$ \Comment{Initial Generated Image}\\

\\
\textit{\textcolor{gray}{\#Plan}}
\State  $\{T_{e_{1}}, T_{e_{2}}, \cdots T_{e_{n}},\} \gets \mathcal{P}\big(I_g, T, \big[E_{1} \cdots E_p\big]\big)$\\

\\
\textit{\textcolor{gray}{\#Edit}}
\State $I_{e_0} \gets I_g$
\For {k = 1, $\cdots$, n-1}
    \State $I_{e_{k+1}} \gets \mathcal{E}(I_{e_k}, T_{e_{k+1}})$\Comment{Intermediate Edited Image}
\EndFor
\State $I_{e_o} \gets I_{e_n}$ \Comment{Final Edited Image}\\

\end{algorithmic}
\end{algorithm}

\section{Ablations:}
\label{supp:ablations}

\begin{figure*}
    \centering
    \includegraphics[width=1.0\linewidth]{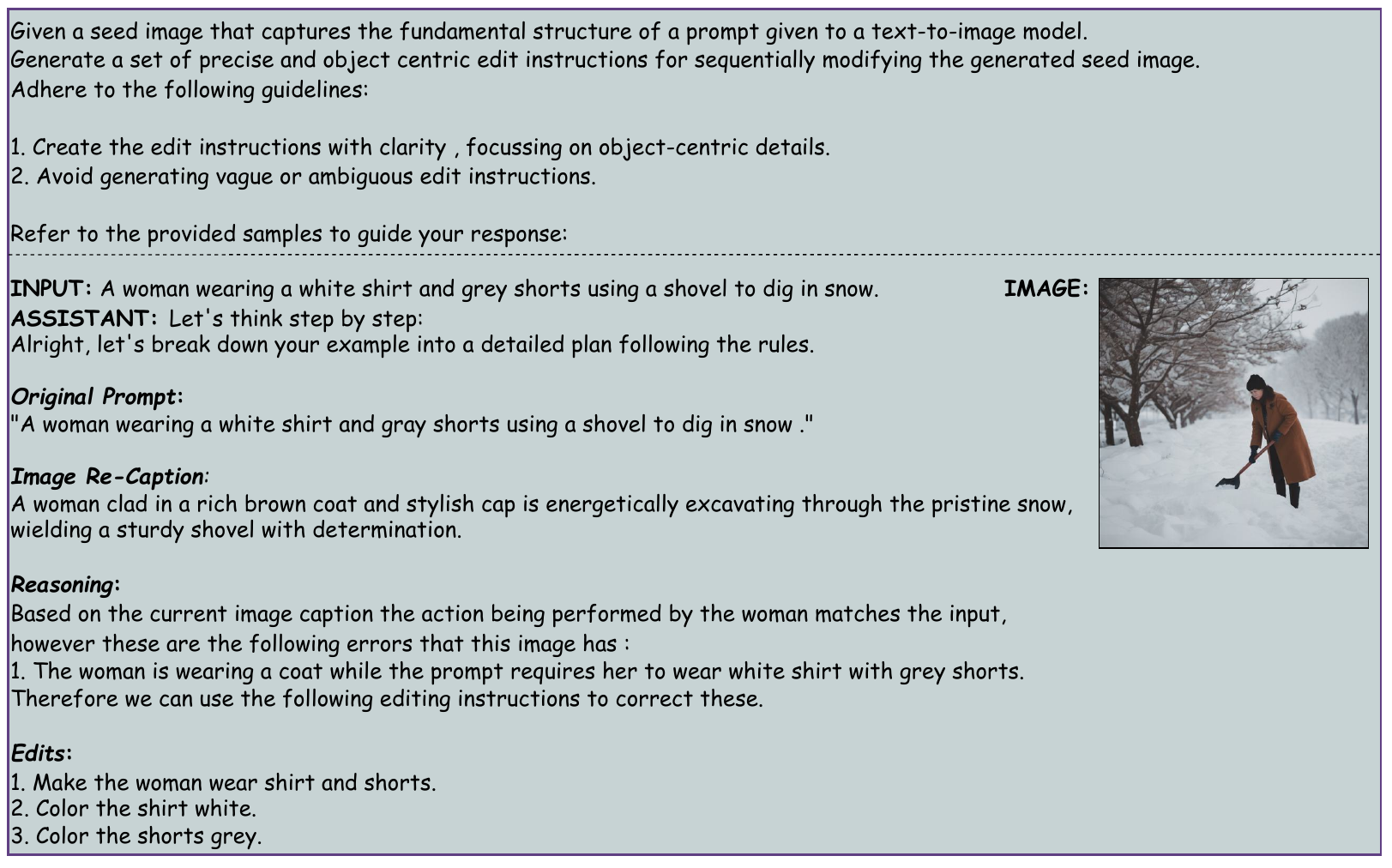}
    \caption{System-prompt and few-shot example format for GraPE$_{naive}$}
    \label{fig:prompt_design_grape_naive}
\end{figure*}

We provide the system prompt and an in-context prompting example in Fig.~\ref{fig:prompt_design_grape_naive} for the GraPE$_{naive}$ pipeline used in section ~\ref{naive_planner_ablation}. The planner follows the system prompt to generate reasoning and editing instructions from the given image and text prompt, without the proposed decomposition into image and text based elements.

\begin{figure*}
    \centering
    \includegraphics[width=1.0\linewidth]{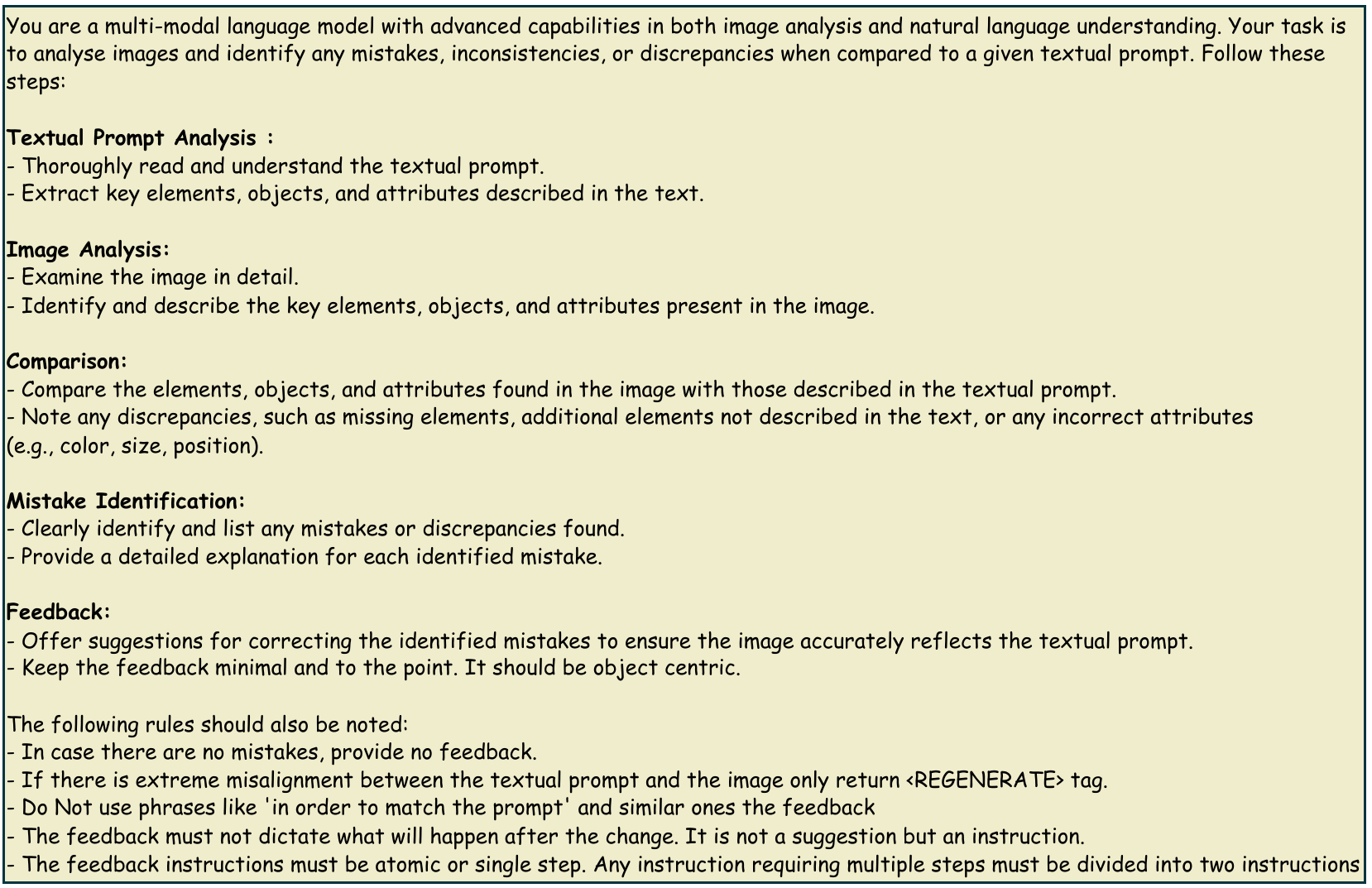}
    \caption{System-prompt used with GraPE's MLLM Planner}
    \label{fig:system_prompt}
\end{figure*}

\begin{figure*}
    \centering
    \includegraphics[width=1.0\linewidth]{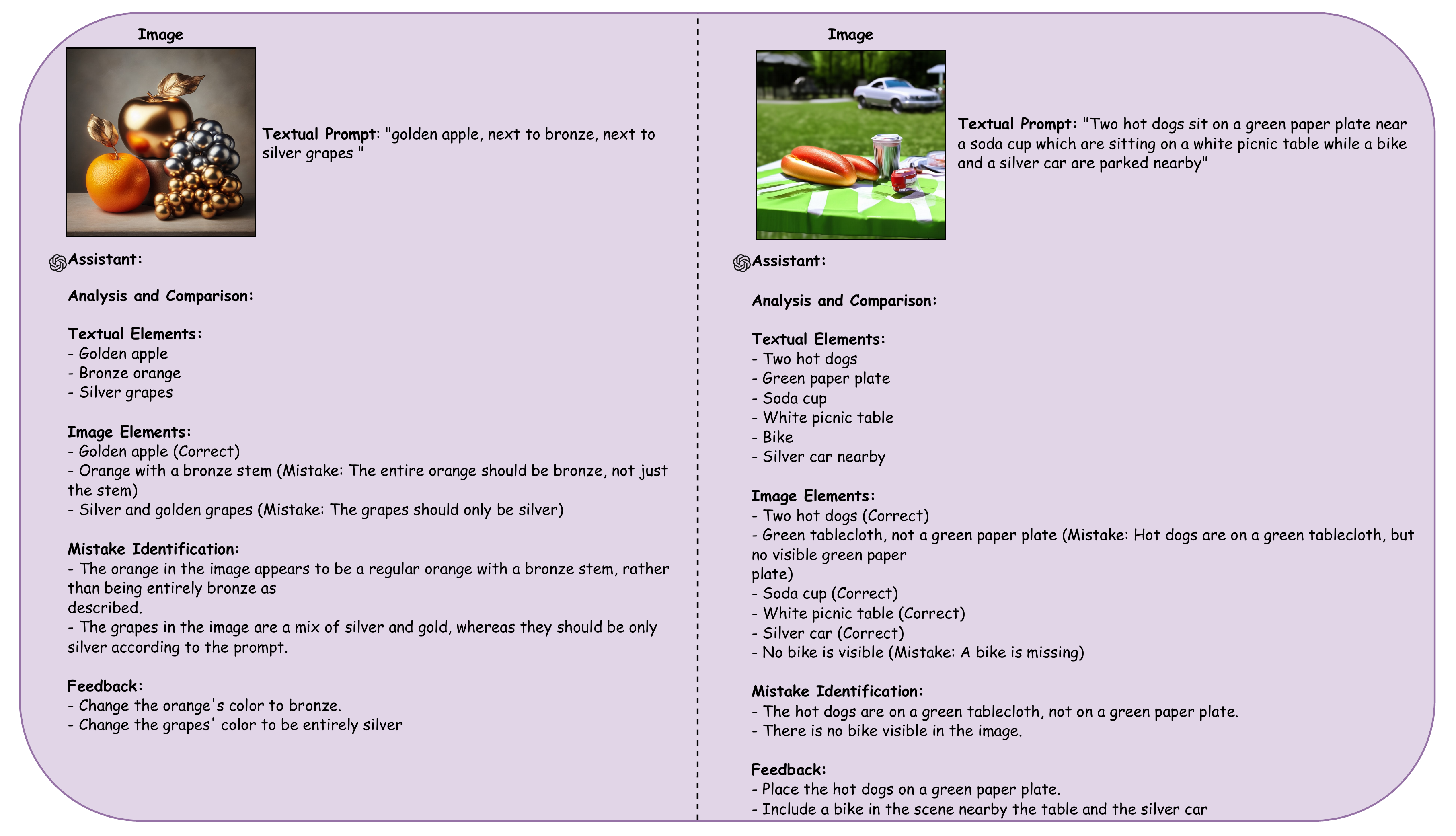}
    \caption{Selective Few shot examples used in GraPE's MLLM Planner}
    \label{fig:few_shot_ex}
\end{figure*}

\section{Additional and Detailed Results:}
\label{supp:more_results}

\paragraph{Comparison with SOTA LLM-Based Approaches}
\label{sec:SLD}
In this section, we extend our comparison to recently proposed SLD framework~\cite{wu2023self} that generates an image from the input prompt, assesses its alignment with the prompt, and performs self-corrections on the inaccuracies in the generated image. Unlike~\algo, SLD operates in image layout-space which is either generated using either LLMs or open-vocabulary object detectors. The corrective steps are defined over object latents within the layout. Table~\ref{tab:rebut_table} shows the performance of SLD when compared to \algo~ on ConceptMix benchmark. \algo~ consistently outperforms SLD in rectifying generated images across models and prompt-complexities. \\
Note on RPG: We find that RPG fails to generate valid plans for over 50\% of the samples in the comparison, primarily due to frequent parsing errors, therefore the numbers reported are on the successfully generated samples only.

\begin{table}[h]
    \centering

    \renewcommand{\arraystretch}{1.5}
    \resizebox{\columnwidth}{!}{
    \begin{tabular}{l|ccc|ccc|ccc|ccc}
        \toprule
        \textbf{Complexity $\rightarrow$} & \multicolumn{3}{c|}{\textbf{K=1}} & \multicolumn{3}{c|}{\textbf{K=3}} & \multicolumn{3}{c|}{\textbf{K=5}} & \multicolumn{3}{c}{\textbf{K=7}} \\ 
        \textbf{Method $\rightarrow$} & Base & SLD & \algo\ & Base & SLD & \algo\  & Base & SLD & \algo\  & Base & SLD & \algo\  \\ 
        \midrule

        SD1.5                           & 0.81 & 0.87 & \textbf{0.89} 
                                        & 0.61 & 0.64 & \textbf{0.75} 
                                        & 0.50 & 0.61 & \textbf{0.69}
                                        & 0.45 & 0.55 & \textbf{0.61} \\ 
        
        StructDiff                      & 0.82 & \textbf{0.90} & \textbf{0.90} 
                                        & 0.61 & 0.66 & \textbf{0.72} 
                                        & 0.54 & 0.64 & \textbf{0.70}
                                        & 0.44& 0.51& \textbf{0.57} \\ 

        SD2.1                           & 0.83 & \textbf{0.87} & \textbf{0.87} 
                                        & 0.64 & 0.68 & \textbf{0.74} 
                                        & 0.58 & 0.63 & \textbf{0.69}
                                        & 0.47& 0.57 & \textbf{0.63} \\ 

        SDXL                            & 0.85 & 0.85 & \textbf{0.88} 
                                        & 0.71 & 0.68 &\textbf{ 0.78} 
                                        & 0.64 & 0.64 & \textbf{0.73} 
                                        & 0.52 & 0.53 & \textbf{0.63}\\

        PixArt                          & 0.81 & 0.87 & \textbf{0.87} 
                                        & 0.67 & 0.68 &\textbf{ 0.72} 
                                        & 0.65 & 0.67 & \textbf{0.74} 
                                        & 0.51 & 0.55 & \textbf{0.63}\\

        DeepFloyd-IF                    & 0.88 & 0.91 & \textbf{0.92} 
                                        & 0.68 & 0.71 & \textbf{0.77} 
                                        & 0.66 & 0.68 & \textbf{0.75}
                                        & 0.58& 0.62 & \textbf{0.66} \\ 

        PlaygroundV2.5                  & 0.91 & 0.91 & \textbf{0.96} 
                                        & 0.74 & 0.73 & \textbf{0.79} 
                                        & 0.66 & 0.66 & \textbf{0.72}
                                        & 0.54& 0.54 & \textbf{0.64} \\

        Dalle3                          & 0.95 & 0.89 & \textbf{0.95} 
                                        & 0.83 & 0.77 &\textbf{ 0.86} 
                                        & 0.81 & 0.77 & \textbf{0.83} 
                                        & 0.73 & 0.65 & \textbf{0.74}\\

        SD3.5 Large                     & 0.93 & 0.90 & \textbf{0.95} 
                                        & 0.81 & 0.80 &\textbf{ 0.82}
                                        & 0.80 & 0.73 & \textbf{0.83} 
                                        & 0.76 & 0.72 & \textbf{0.78}\\

        Flux                            & 0.90 & 0.90 & \textbf{0.92} 
                                        & 0.82 & 0.82 &\textbf{ 0.87} 
                                        & 0.79 & 0.73 & \textbf{0.81} 
                                        & \textbf{0.78} & 0.68 & \textbf{0.78}\\

        LMD                             & 0.86 & 0.85 & \textbf{0.87} 
                                        & 0.71 & 0.71 &\textbf{ 0.77} 
                                        & 0.64 & 0.64 & \textbf{0.73} 
                                        & 0.59 & 0.59 & \textbf{0.67}\\
                                        
        RAG                             & 0.81 & 0.82 & \textbf{0.87}
                                        & 0.67 & 0.71 &\textbf{0.72} 
                                        & 0.66 & 0.68 & \textbf{0.72} 
                                        & 0.52 & 0.56 & \textbf{0.63}\\

        RPG                             & 0.70 & 0.73 & \textbf{0.85}
                                        & 0.69 & 0.68 &\textbf{0.72} 
                                        & 0.58 & 0.64 & \textbf{0.69} 
                                        & 0.39 &\textbf{ 0.54} & 0.47\\

        \bottomrule
    \end{tabular}}
    \vspace{-10pt}
    \caption{Comparison of performance between SLD and \algo~ when applied over Base images as post-hoc correction.}
    \label{tab:rebut_table}
    \vspace{-0.3cm}
\end{table}



\paragraph{Table for T2I-Bench and Flickr-Bench} Table~\ref{table:t2i_compbench_bench_supp},~\ref{tab:flickr_bench_detailed} presents the absolute values of \algo\ with PixEdit and AURORA~\cite{krojer2024learningactionreasoningcentricimage} as editing models, for the graphs in Fig.~\ref{fig:flickr_bench_combined}.

\paragraph{Results on ConceptMix Benchmark using \algo\ with AURORA} Table~\ref{tab:conceptmix_detailed_aurora} presents the numbers on ConceptMix Dataset using \algo\ with AURORA editing model, this table complements Table~\ref{table:concept_mix_pixedit}, where PixEdit was used as the editing model.

\paragraph{Performance on Editing Benchmarks}
We employ two established benchmarks to evaluate and compare the editing performance of PixEdit with other models. The first benchmark, AURORA-BENCH, consists of image-edit instruction pairs sourced from eight distinct datasets. This benchmark enables the assessment of the discriminative editing capabilities of models across a wide variety of images. Performance is measured using an automated metric called \textbf{DiscEdit}~\cite{krojer2024learningactionreasoningcentricimage}. The second benchmark we evaluate is the MagicBrush Test-set, which comprises 1K editing turns, including both single-turn and multi-turn edits. For quantitative analysis on this benchmark, we utilize standard metrics such as L1, L2, and CLIP/DINO similarity scores, providing a comprehensive evaluation of the editing performance. The results in Table~\ref{tab:performance on AURORA-BENCH} indicates strong discriminative capability in PixEdit compared to baselines. Table~\ref{tab:MB-TEST set performance} shows that PixEdit is great at single turn edits and follows closely with baselines in multi-turn edits.

\begin{table}[h]
    \centering
    \resizebox{\columnwidth}{!}{
    \begin{tabular}{lccccc}
        \toprule
        \textbf{Model} & \textbf{WhatsUp} & \textbf{Something} & \textbf{AG} & \textbf{Kubric} & \textbf{CLEVR} \\
        \midrule
        MagicBrush & 0.472 & 0.371 & 0.477 & 0.392 & 0.400 \\
        AURORA & 0.565 & 0.548 & 0.583 & \textbf{0.592} & 0.450 \\
        PixEdit &\textbf{0.566} & \textbf{0.613} & \textbf{0.606} & 0.400 & \textbf{0.600} \\
        \bottomrule
    \end{tabular}
    }
    \caption{Performance comparison of PixEdit with other models on AURORA-BENCH using \textit{DiscEdit} computed as specified in~\cite{krojer2024learningactionreasoningcentricimage}. Note: The baseline scores are taken from ~\cite{krojer2024learningactionreasoningcentricimage}}
    \label{tab:performance on AURORA-BENCH}
\end{table}

\begin{table*}[h!]
    \centering
    \begin{tabular}{lcccccc}
        \toprule
        \textbf{Settings} & \textbf{Methods} & \textbf{L1$\downarrow$} & \textbf{L2$\downarrow$} & \textbf{CLIP-I$\uparrow$} & \textbf{DINO$\uparrow$} & \textbf{CLIP-T$\uparrow$} \\
        \midrule
        \multirow{3}{*}{\textbf{Single-turn}} 
        & MagicBrush & 0.0788 & 0.0274 & 0.8978 & 0.8313 & 0.2973 \\
        & AURORA & 0.0754 & 0.0270 & 0.9105 & 0.8594 & \textbf{0.2981} \\
        & PixEdit & \textbf{0.0719} & 0.0278 & 0.9082 & \textbf{0.8661} & 0.2946 \\
        \midrule
        \multirow{3}{*}{\textbf{Multi-Turn}} 
        & MagicBrush & 0.0921 & 0.0327 & 0.8777 & 0.7996 & \textbf{0.3020} \\
        & AURORA & 0.0904 & 0.0334 & \textbf{0.8887} & \textbf{0.8272} & 0.3005 \\
        & PixEdit & \textbf{0.0892} & 0.0357 & 0.8729 & 0.8068 & 0.2947 \\
        \bottomrule
    \end{tabular}
    \caption{Comparison of methods across Single-turn and Multi-turn settings on Magic-Brush Test set. \textbf{Note}: The numbers were reproduced using the publicly available codebases of the respective baselines with the same random seed.}
    \label{tab:MB-TEST set performance}
\end{table*}

\paragraph{Cost-Analysis}
We quantify the additional cost incurred by GraPE primarily through the planning phase, which leverages GPT-4o for generating precise, atomic edit instructions. Our measurements indicate that this planning stage costs approximately \$0.013 per image, amounting to roughly \$52 for processing the entire ConceptMix benchmark. We note that knowledge distillation methods described in OmniEdit~\cite{wei2024omnieditbuildingimageediting} and LLaVa~\cite{liu2023visualinstructiontuning} could help by transferring GPT-4o level capabilities to smaller open-source models, potentially reducing inference costs. The modular design of the framework enables a flexible trade-off between performance and compute cost.

\paragraph{Runtime-Comparison}
We carry out runtime evaluation of \algo~on an A100 GPU. We find that the planning phase requires an average of 20 seconds per image, with the editing phase adding approximately 3 seconds for each atomic edits—given that we perform an average of 2.53 edit steps, calculated across benchmarks and models. This is in contrast to other post-hoc correction methods, such as SLD and RAG, which average at 52.5 and 55 seconds per image, respectively. Considering that the average base generation time for the top 5 performing text-to-image models is 20 seconds (approx), our approach effectively leverages iterative refinement to bridge the performance gap between weaker and stronger models without incurring excessive computational delays.

\section{Implementation Details:}
\label{supp:implementation_details}

\subsection{Prompts and Few-shot Examples:}
Figure~\ref{fig:system_prompt} and Figure~\ref{fig:few_shot_ex} illustrate the full system prompt and a selection of few-shot examples employed in the MLLM-based planner for \algo.
\subsection{Hyperparameters}
\paragraph{Generation} To generate images using various diffusion models, we adhere to the default hyperparameters specific to each model, such as the number of inference steps and the sampling method. All images are generated with a fixed random seed of 0 to ensure reproducibility.

\paragraph{Planning} We utilize GPT-4o with a sampling temperature set to 0, ensuring deterministic outputs for both planning and Visual Question Answering (VQA) tasks.

\paragraph{Editing} For image editing, we use AURORA with 50 inference steps and its default sampler as described in its paper. Additionally, we employ 14 inference steps with DPM-Solver++ in combination with the PixEdit model to generate edited images, ensuring reproducibility by using a fixed random seed of 0.

\subsection{PixEdit:}
\label{supp:pixedit_implementation_details}
This section details the implementation of PixEdit model. We chose PixArt-sigma~\cite{chen2024pixartsigmaweaktostrongtrainingdiffusion} as our backbone T2I model which we convert into an editing checkpoint. We follow similar training strategy as used in ~\citep{brooks2023instructpix2pixlearningfollowimage, krojer2024learningactionreasoningcentricimage} i.e adding additional channels in the initial convolution layer of the diffusion model and use the randomly sampled noise concatenated with input image as input to de-noise into edited-image. We started with the SEED-EDIT~\cite{ge2024seeddataedittechnicalreporthybrid} dataset for initial pre-training stage. We skip the automatically generated data and use the 147K sample subset of this dataset that are either procured from the web or manually created to ensure high quality pre-training data. The model is trained for 32K steps with an effective batch size of 256 at this stage. We further fine-tune this model for 42K steps on the mixture of editing dataset proposed in~\cite{krojer2024learningactionreasoningcentricimage} to create PixEdit.

\subsubsection{Analysis of Improvement in PixEdit}
PixEdit introduces two significant advancements over AURORA:\\
\textbf{(1) Upgraded Backbone Diffusion Model:} Transitioning from SD1.5 to PixArt-Sigma.\\
\textbf{(2) Refined Pre-Training Data}: Replacing the Instruct-Pix2Pix dataset with the real-only subset of the Seed-Edit dataset for pre-training.\\
To isolate the impact of change (2), we compare PixEdit with SD1.5 adapted for editing using the same Seed-Edit dataset and evaluate their performance on the $K=7$ subset of the ConceptMix benchmark. Results in Table~\ref{table:k=7_seed_vs_pixedit} show that SD1.5 adapted with Seed-Edit data, and subsequently fine-tuned using the Aurora data-mixture, performs slightly better than AURORA alone on average. This highlights the advantage of incorporating real-world, complex pre-training data.

PixEdit, leveraging its more powerful PixArt-Sigma backbone and a compositional text-encoder, achieves the best performance on average. This underscores the benefits of combining an advanced diffusion model with improved text-image alignment for editing tasks.

\begin{table*}[h!]
    \centering
    \begin{tabular}{l |r r r}
    \toprule
    \multirow{2}{*}{Method} & \multicolumn{3}{c}{Concept K=7}\\ 
    \cline{2-4}
    & \multicolumn{1}{c}{\algo\textsubscript{AURORA}}& \multicolumn{1}{c}{\algo\textsubscript{SD1.5$^{*}$}} & \multicolumn{1}{c}{\algo\textsubscript{PixEdit}}\\ 
                                 
    \midrule                                                                        
    Stable-Diffusion v1.5 \citep{Rombach_2022_CVPR}                                     
                                                                                        & $0.598$ \color{gray}$_{ \pm 0.003}$
                                                                                        & $\textbf{0.618}$ \color{gray}$_{ \pm 0.000}$    
                                                                                        & $0.610$ \color{gray}$_{ \pm 0.005}$\\
                                
    Structure Diffusion \citep{feng2023trainingfreestructureddiffusionguidance}        
                                                                                        & $\textbf{0.612}$ \color{gray}$_{ \pm 0.003}$
                                                                                        & $0.596$ \color{gray}$_{ \pm 0.005}$
                                                                                        & $0.571$ \color{gray}$_{ \pm 0.002}$ \\
                                
    Stable Diffusion v2.1 \citep{Rombach_2022_CVPR}                                
                                                                                        & $0.613$ \color{gray}$_{ \pm 0.009}$
                                                                                        & $0.588$ \color{gray}$_{ \pm 0.003}$  
                                                                                        & $\textbf{0.626}$ \color{gray}$_{ \pm 0.002}$\\
    
    LMD \citep{lian2024llmgroundeddiffusionenhancingprompt}                             
                                                                                        & $0.625$ \color{gray}$_{ \pm 0.005}$
                                                                                        & $0.649$ \color{gray}$_{ \pm 0.004}$  
                                                                                        & $\textbf{0.668}$ \color{gray}$_{ \pm 0.009}$\\
    
    SD-XL \citep{podell2023sdxlimprovinglatentdiffusion}                               
                                                                                       & $0.624$ \color{gray}$_{ \pm 0.010}$
                                                                                       & $0.615$ \color{gray}$_{ \pm 0.008}$   
                                                                                       & $\textbf{0.628}$ \color{gray}$_{ \pm 0.002}$\\
    
    PixArt-$\alpha$ \citep{chen2023pixartalphafasttrainingdiffusion}                     
                                                                                        & $0.624$ \color{gray}$_{ \pm 0.006}$
                                                                                        & $0.614$ \color{gray}$_{ \pm 0.006}$
                                                                                        & $\textbf{0.625}$ \color{gray}$_{ \pm 0.002}$\\
    
    DeepFloyd IF \citep{DeepFloydIF}                                                     
                                                                                      & $0.638$ \color{gray}$_{ \pm 0.006}$
                                                                                       & $\textbf{0.672}$ \color{gray}$_{ \pm 0.007}$ 
                                                                                       & $0.662$ \color{gray}$_{ \pm 0.006}$\\
    
    PlaygroundV2.5 \citep{li2024playgroundv25insightsenhancing}                          
                                                                                       & $0.611$ \color{gray}$_{ \pm 0.002}$
                                                                                        & $\textbf{0.722}$ \color{gray}$_{ \pm 0.004}$ 
                                                                                        & $0.640$ \color{gray}$_{ \pm 0.005}$\\
    
    Dalle3 \citep{betker2023improving}                                                   
                                                                                    & $0.718$ \color{gray}$_{ \pm 0.006}$
                                                                                        & $0.719$ \color{gray}$_{ \pm 0.007}$   
                                                                                        & $\textbf{0.737}$ \color{gray}$_{ \pm 0.004}$\\
    
    Stable Diffusion v3.5 Large \citep{sd3_5_dummy}                                 
                                                                                    & $0.765$ \color{gray}$_{ \pm 0.003}$
                                                                                       & $0.761$ \color{gray}$_{ \pm 0.007}$  
                                                                                       & $\textbf{0.784}$ \color{gray}$_{ \pm 0.005}$\\  

    Flux-schnell \citep{flux}                                 
                                                                                    & $0.748$ \color{gray}$_{ \pm 0.007}$
                                                                                    & $0.785$ \color{gray}$_{ \pm 0.006}$  
                                                                                    & $\textbf{0.779}$ \color{gray}$_{ \pm 0.004}$\\  

    RAG \citep{flux}                                 
                                                                                    & $0.604$ \color{gray}$_{ \pm 0.004}$
                                                                                    & $0.602$ \color{gray}$_{ \pm 0.003}$  
                                                                                    & $\textbf{0.628}$ \color{gray}$_{ \pm 0.001}$\\  

    RPG \citep{flux}                                 
                                                                                    & $\textbf{0.505}$ \color{gray}$_{ \pm 0.003}$
                                                                                    & $0.473$ \color{gray}$_{ \pm 0.005}$  
                                                                                    & $0.467$ \color{gray}$_{ \pm 0.005}$\\  
    
    \bottomrule
    \end{tabular}
    \caption{GPT-QA scores using \algo\ with different editing models SD1.5$^{*}$ vs AURORA vs PixEdit. Note \algo\textsubscript{SD1.5$^{*}$} represents the editing model based on SD1.5 trained with refined pre-training data and fine-tuned on Aurora data-mixture.}
    \label{table:k=7_seed_vs_pixedit}
    \end{table*}

\section{Additional Qualitative Examples:}
\label{supp:qualitative}

\paragraph{Errors due to Planner vs Editing Model}
We try to understand the broader context, consider a hypothetical set of 100 plans (randomly sampled from the human study on MLLM planners). We look at the breakup of these 100 plans and how the planner and editing-model fare on them. $\sim74.5\%$ of the plans are deemed to have generation errors, $\sim55.5\%$ ($41.3$ plans in absolute terms) of which are completely corrected at the planner level (considering GPT-4o) indicating a strong alignment between the generated plan and the text prompt. We then note that about one-third of these ($\sim30.5\%$ or $12.5$ images) are successfully edited to produce completely correct final images, showcasing the editing model's ability to translate plans into precise visual outputs. The remaining two-thirds of the aligned plans ($\sim69.5\%$ or $28.8$ images) fall short, with errors entirely attributable to the editing model. This underscores that while the planner demonstrates robust performance in generating accurate and contextually aligned plans, the editing model remains the predominant source of errors.

\paragraph{Editing Errors in Perfect Plans}
The subset of 46 edit plans showing full improvement as per the majority of human annotators chosen for qualitative error analysis predominantly involved plans requiring "add" actions, with relatively fewer plans focusing on "remove" or "modify" actions. This can also be seen from the breakdown of the corresponding final edited images on the basis of mentioned actions. This distribution reflects the natural bias of generative models which often fail to identify certain textual elements and the image generated then has errors which can be fixed by "add" edit actions.

\paragraph{Plans showing no or partial improvement}
Finally, considering the plans which show no/partial improvement we note that distinct patterns emerge providing further insights into the limitations of the planner. The plans with no improvement are primarily empty plans generated by the planner, where it fails to propose any actionable edits to address the misalignment between the text prompt and the image. On the other hand, the plans with partial improvement typically exhibit incomplete details, where some, but not all, elements of the prompt are addressed. A small subset of these also includes hallucinations, where the planner mistakenly assumes the presence of an object in the image that does not actually exist. These observations highlight areas where the planner could be enhanced, particularly in terms of improving its ability to identify actionable edits and reducing cases of over- or under-specification in the generated plans.

\begin{table*}[h!]
\label{table:concept_mix}
\setlength{\tabcolsep}{6pt}
\renewcommand{\arraystretch}{1.2}
\centering
\resizebox{\textwidth}{!}{
\begin{tabular}{l |r r|r r|r r|r r}
\toprule
\multirow{2}{*}{Method} & \multicolumn{2}{c |}{Concept K=1} & \multicolumn{2}{c|}{Concept K=3} & \multicolumn{2}{c|}{Concept K=5} & \multicolumn{2}{c}{Concept K=7}\\ 
\cline{2-9}
& \multicolumn{1}{c}{ Base } & \multicolumn{1}{c|}{\algo\textsubscript{AURORA}} & \multicolumn{1}{c}{Base} & \multicolumn{1}{c|}{\algo\textsubscript{AURORA}} & \multicolumn{1}{c}{Base} & \multicolumn{1}{c|}{\algo\textsubscript{AURORA}} & \multicolumn{1}{c}{Base} & \multicolumn{1}{c}{\algo\textsubscript{AURORA}}\\ 
                             
\midrule                
\midrule
Stable-Diffusion v1.5 \cite{Rombach_2022_CVPR}       
                                                        & $0.808$ \color{gray}$_{ \pm 0.009}$             
                                                        & $\textbf{0.863}$ \color{gray}$_{ \pm 0.005}$ 
                                                        & $0.606$ \color{gray}$_{ \pm 0.018}$  
                                                        & $\textbf{0.717}$ \color{gray}$_{ \pm 0.024}$ 
                                                        & $0.497$ \color{gray}$_{ \pm 0.010}$ 
                                                        & $\textbf{0.688}$ \color{gray}$_{ \pm 0.010}$ 
                                                        & $0.450$ \color{gray}$_{ \pm 0.005}$ 
                                                        & $\textbf{0.598}$ \color{gray}$_{ \pm 0.003}$ \\
                            
Structure Diffusion \cite{feng2023trainingfreestructureddiffusionguidance}           & $0.823$ \color{gray}$_{ \pm 0.002}$      
                                                                                    & $\textbf{0.895}$ \color{gray}$_{ \pm 0.004}$ 
                                                                                   & $0.606$ \color{gray}$_{ \pm 0.002}$ 
                                                                                    & $\textbf{0.734}$ \color{gray}$_{ \pm 0.021}$ 
                                                                                    & $0.542$ \color{gray}$_{ \pm 0.014}$ 
                                                                                    & $\textbf{0.698}$ \color{gray}$_{ \pm 0.015}$
                                                                                    & $0.447$ \color{gray}$_{ \pm 0.001}$ 
                                                                                    & $\textbf{0.612}$ \color{gray}$_{ \pm 0.003}$ \\
                            
Stable Diffusion v2.1 \cite{Rombach_2022_CVPR}      
                                                                                & $0.833$ \color{gray}$_{ \pm 0.002}$ 
                                                                               & $\textbf{0.875}$ \color{gray}$_{ \pm 0.014}$ 
                                                                                & $0.639$ \color{gray}$_{ \pm 0.014}$ 
                                                                                & $\textbf{0.719}$ \color{gray}$_{ \pm 0.012}$
                                                                                & $0.579$ \color{gray}$_{ \pm 0.012}$ 
                                                                                & $\textbf{0.694}$ \color{gray}$_{ \pm 0.009}$
                                                                                & $0.466$ \color{gray}$_{ \pm 0.002}$ 
                                                                                & $\textbf{0.613}$ \color{gray}$_{ \pm 0.009}$ \\

LMD \cite{lian2024llmgroundeddiffusionenhancingprompt}                         
                                                                            & $0.855$ \color{gray}$_{ \pm 0.004}$ 
                                                                           & $\textbf{0.907}$ \color{gray}$_{ \pm 0.008}$ 
                                                                            & $0.711$ \color{gray}$_{ \pm 0.008}$ 
                                                                           & $\textbf{0.765}$ \color{gray}$_{ \pm 0.005}$
                                                                            & $0.643$ \color{gray}$_{ \pm 0.011}$ 
                                                                            & $\textbf{0.721}$ \color{gray}$_{ \pm 0.012}$
                                                                            & $0.591$ \color{gray}$_{ \pm 0.002}$ 
                                                                            & $\textbf{0.625}$ \color{gray}$_{ \pm 0.005}$ \\

SD-XL \cite{podell2023sdxlimprovinglatentdiffusion}                       
                                                                        & $0.848$ \color{gray}$_{ \pm 0.010}$  
                                                                        & $\textbf{0.893}$ \color{gray}$_{ \pm 0.006}$ 
                                                                         & $0.708$ \color{gray}$_{ \pm 0.018}$ 
                                                                        & $\textbf{0.740}$ \color{gray}$_{ \pm 0.013}$ 
                                                                        & $0.635$ \color{gray}$_{ \pm 0.014}$ 
                                                                        & $\textbf{0.692}$ \color{gray}$_{ \pm 0.018}$
                                                                        & $0.520$ \color{gray}$_{ \pm 0.003}$
                                                                        & $\textbf{0.624}$ \color{gray}$_{ \pm 0.010}$\\

PixArt-$\alpha$ \cite{chen2023pixartalphafasttrainingdiffusion}                
                                                                    & $0.813$ \color{gray}$_{ \pm 0.010}$ 
                                                                    & $\textbf{0.883}$ \color{gray}$_{ \pm 0.008}$ 
                                                                    & $0.668$ \color{gray}$_{ \pm 0.011}$
                                                                    & $\textbf{0.717}$ \color{gray}$_{ \pm 0.012}$ 
                                                                    & $0.649$ \color{gray}$_{ \pm 0.011}$ 
                                                                    & $\textbf{0.711}$ \color{gray}$_{ \pm 0.022}$
                                                                    & $0.507$ \color{gray}$_{ \pm 0.001}$
                                                                    & $\textbf{0.624}$ \color{gray}$_{ \pm 0.006}$\\

DeepFloyd IF \cite{DeepFloydIF}                
                                                                & $0.883$ \color{gray}$_{ \pm 0.009}$
                                                                & $\textbf{0.898}$ \color{gray}$_{ \pm 0.005}$ 
                                                                & $0.680$ \color{gray}$_{ \pm 0.016}$
                                                                & $\textbf{0.735}$ \color{gray}$_{ \pm 0.016}$
                                                                & $0.663$ \color{gray}$_{ \pm 0.014}$
                                                                & $\textbf{0.713}$ \color{gray}$_{ \pm 0.007}$
                                                                & $0.583$ \color{gray}$_{ \pm 0.002}$ 
                                                                & $\textbf{0.638}$ \color{gray}$_{ \pm 0.006}$\\

PlaygroundV2.5 \cite{li2024playgroundv25insightsenhancing} & $0.908$ \color{gray}$_{ \pm 0.010}$ 
                                                            & $\textbf{0.945}$ \color{gray}$_{ \pm 0.004}$ 
                                                            & $0.737$ \color{gray}$_{ \pm 0.023}$
                                                            & $\textbf{0.783}$ \color{gray}$_{ \pm 0.012}$ 
                                                            & $0.658$ \color{gray}$_{ \pm 0.015}$
                                                            & $\textbf{0.709}$ \color{gray}$_{ \pm 0.006}$
                                                            & $0.540$ \color{gray}$_{ \pm 0.003}$ 
                                                            & $\textbf{0.611}$ \color{gray}$_{ \pm 0.002}$\\

Dalle3 \cite{betker2023improving}                       & $0.947$ \color{gray}$_{ \pm 0.002}$ 
                                                        & $\textbf{0.950}$ \color{gray}$_{ \pm 0.004}$ 
                                                        & $\textbf{0.832}$ \color{gray}$_{ \pm 0.012}$  
                                                        & $0.809$ \color{gray}$_{ \pm 0.007}$ 
                                                        & $\textbf{0.812}$ \color{gray}$_{ \pm 0.014}$ 
                                                        & $0.783$ \color{gray}$_{ \pm 0.015}$
                                                        & $\textbf{0.728}$ \color{gray}$_{ \pm 0.006}$ 
                                                        & $0.718$ \color{gray}$_{ \pm 0.006}$\\

Stable Diffusion v3.5 Large \cite{sd3_5_dummy}       & $\textbf{0.927}$ \color{gray}$_{ \pm 0.005}$  
                                                     & $\textbf{0.927}$ \color{gray}$_{ \pm 0.002}$ 
                                                     & $\textbf{0.815}$ \color{gray}$_{ \pm 0.002}$  
                                                    & $0.812$ \color{gray}$_{ \pm 0.004}$ 
                                                     & $\textbf{0.803}$ \color{gray}$_{ \pm 0.003}$ 
                                                     & $0.792$ \color{gray}$_{ \pm 0.002}$
                                                     & $0.759$ \color{gray}$_{ \pm 0.004}$ 
                                                     & $\textbf{0.765}$ \color{gray}$_{ \pm 0.003}$\\

Flux-schnell \citep{flux}                                          & $0.902$ \color{gray}$_{ \pm 0.002}$ 
                                                                                    & $\textbf{0.940}$ \color{gray}$_{\pm 0.004}$
                                                                                    & $0.820$ \color{gray}$_{ \pm 0.003}$ 
                                                                                    & $\textbf{0.856}$ \color{gray}$_{ \pm 0.001}$ 
                                                                                    & $0.786$ \color{gray}$_{ \pm 0.005}$ 
                                                                                    & $\textbf{0.781}$ \color{gray}$_{ \pm 0.002}$ 
                                                                                    & $0.775$ \color{gray}$_{ \pm 0.004}$ 
                                                                                    & $\textbf{0.748}$ \color{gray}$_{ \pm 0.007}$ \\

    RAG \citep{chen2024regionaware}                        & $0.815$ \color{gray}$_{ \pm 0.009}$ 
                                                                        & $\textbf{0.865}$ \color{gray}$_{\pm 0.005}$
                                                                        & $0.668$ \color{gray}$_{ \pm 0.007}$ 
                                                                        & $\textbf{0.718}$ \color{gray}$_{ \pm 0.005}$ 
                                                                        & $0.665$ \color{gray}$_{ \pm 0.002}$ 
                                                                        & $\textbf{0.687}$ \color{gray}$_{ \pm 0.007}$ 
                                                                        & $0.520$ \color{gray}$_{ \pm 0.003}$ 
                                                                        & $\textbf{0.604}$ \color{gray}$_{ \pm 0.004}$ \\

    RPG \citep{yang2024mastering}                                           & $0.696$ \color{gray}$_{ \pm 0.015}$ 
                                                                            & $\textbf{0.831}$ \color{gray}$_{\pm 0.001}$
                                                                            & $0.694$ \color{gray}$_{ \pm 0.002}$ 
                                                                            & $\textbf{0.756}$ \color{gray}$_{ \pm 0.005}$ 
                                                                            & $0.583$ \color{gray}$_{ \pm 0.002}$ 
                                                                            & $\textbf{0.688}$ \color{gray}$_{ \pm 0.007}$ 
                                                                            & $0.388$ \color{gray}$_{ \pm 0.007}$ 
                                                                            & $\textbf{0.505}$ \color{gray}$_{ \pm 0.003}$ \\

\bottomrule
\end{tabular}
}
\caption{Results on Concept-mix benchmark - \algo\textsubscript{AURORA}, Complementary to Table~\ref{table:concept_mix_pixedit}}
\label{tab:conceptmix_detailed_aurora}
\end{table*}

\begin{table*}[h!]
    \setlength{\tabcolsep}{6pt}
    \renewcommand{\arraystretch}{1.2}

    \centering
    \resizebox{\textwidth}{!}{
    \begin{tabular}{l|cc|cc|cc}
        \toprule
        \multirow{2}{*}{Method} & \multicolumn{2}{c|}{Base} & \multicolumn{2}{c|}{\algo\textsubscript{AURORA}} & \multicolumn{2}{c}{\algo\textsubscript{PixEdit}}\\
        \cline{2-7}
        & \multicolumn{1}{c}{DSG (w/o dep)} & \multicolumn{1}{c|}{DSG}  & \multicolumn{1}{c}{DSG (w/o dep)} & \multicolumn{1}{c|}{DSG}  & \multicolumn{1}{c}{DSG (w/o dep)} & \multicolumn{1}{c}{DSG} \\
        \midrule

        Stable-Diffusion v1.5 \cite{Rombach_2022_CVPR}                                      & $0.679$ \color{gray}$_{ \pm 0.006}$ 	& $0.654$ \color{gray}$_{ \pm 0.007}$ 
                                                                                            & $0.758$ \color{gray}$_{ \pm 0.008}$	& $0.746$ \color{gray}$_{ \pm 0.009}$  
                                                                                            & $\textbf{0.788}$ \color{gray}$_{ \pm 0.006}$	& $\textbf{0.769}$ \color{gray}$_{ \pm 0.006}$ \\
                                                                                            
        Structure Diffusion \cite{feng2023trainingfreestructureddiffusionguidance}          & $0.714$ \color{gray}$_{ \pm 0.004}$ 	& $0.697$ \color{gray}$_{ \pm 0.004}$ 
                                                                                           & $0.760$ \color{gray}$_{ \pm 0.017}$	& $0.741$ \color{gray}$_{ \pm 0.014}$
                                                                                            & $\textbf{0.794}$ \color{gray}$_{ \pm 0.006}$ 	& $\textbf{0.777}$ \color{gray}$_{ \pm 0.008}$ \\
                                                                                            
        Stable Diffusion v2.1 \cite{Rombach_2022_CVPR}                                      & $0.701$ \color{gray}$_{ \pm 0.001}$ 	& $0.688$ \color{gray}$_{ \pm 0.000}$ 
                                                                                            & $0.757$ \color{gray}$_{ \pm 0.012}$	& $0.742$ \color{gray}$_{ \pm 0.014}$           
                                                                                            & $\textbf{0.787}$ \color{gray}$_{ \pm 0.002}$	& $\textbf{0.772}$ \color{gray}$_{ \pm 0.004}$ \\
                                                                                            
        LMD \cite{lian2024llmgroundeddiffusionenhancingprompt}                              & $0.782$ \color{gray}$_{ \pm 0.001}$ 	& $0.777$ \color{gray}$_{ \pm 0.002}$ 
                                                                                            & $0.805$ \color{gray}$_{ \pm 0.005}$	& $0.790$ \color{gray}$_{ \pm 0.006}$
                                                                                            & $\textbf{0.836}$ \color{gray}$_{ \pm 0.008}$	& $\textbf{0.826}$ \color{gray}$_{ \pm 0.008}$ \\
                                                                                            
        SD-XL \cite{podell2023sdxlimprovinglatentdiffusion}                                 & $0.805$ \color{gray}$_{ \pm 0.002}$ 	& $0.795$ \color{gray}$_{ \pm 0.001}$ 
                                                                                            & $\textbf{0.844}$ \color{gray}$_{ \pm 0.004}$	& $\textbf{0.829}$ \color{gray}$_{ \pm 0.005}$        
                                                                                            & $0.839$ \color{gray}$_{ \pm 0.003}$ 	& $0.828$ \color{gray}$_{ \pm 0.002}$ \\
                                                                                            
        PixArt-$\alpha$ \cite{chen2023pixartalphafasttrainingdiffusion}                     & $0.710$ \color{gray}$_{ \pm 0.004}$ 	& $0.699$ \color{gray}$_{ \pm 0.006}$ 
                                                                                            & $0.758$ \color{gray}$_{ \pm 0.006}$	& $0.744$ \color{gray}$_{ \pm 0.009}$                   
                                                                                            & $\textbf{0.796}$ \color{gray}$_{ \pm 0.006}$ 	& $\textbf{0.788}$ \color{gray}$_{ \pm 0.006}$ \\
                                                                                            
        DeepFloyd IF \cite{DeepFloydIF}                                                     & $0.822$ \color{gray}$_{ \pm 0.002}$ 	& $0.811$ \color{gray}$_{ \pm 0.004}$ 
                                                                                            & $0.840$ \color{gray}$_{ \pm 0.008}$	& $0.819$ \color{gray}$_{ \pm 0.011}$                   
                                                                                            & $\textbf{0.852}$ \color{gray}$_{ \pm 0.001}$ 	& $\textbf{0.840}$ \color{gray}$_{ \pm 0.001}$ \\
                                                                                            
        PlaygroundV2.5 \cite{li2024playgroundv25insightsenhancing}                          & $0.788$ \color{gray}$_{ \pm 0.002}$ 	& $0.775$ \color{gray}$_{ \pm 0.004}$ 
                                                                                            & $0.821$ \color{gray}$_{ \pm 0.011}$	& $0.801$ \color{gray}$_{ \pm 0.011}$                  
                                                                                            & $\textbf{0.825}$ \color{gray}$_{ \pm 0.006}$ 	& $\textbf{0.808}$ \color{gray}$_{ \pm 0.006}$ \\
                                                                                            
        Dalle3 \cite{betker2023improving}                                                   & $0.932$ \color{gray}$_{ \pm 0.003}$ 	& $0.924$ \color{gray}$_{ \pm 0.003}$
                                                                                            & $0.913$ \color{gray}$_{ \pm 0.014}$	& $0.910$ \color{gray}$_{ \pm 0.015}$                  
                                                                                            & $\textbf{0.934}$ \color{gray}$_{ \pm 0.003}$ 	& $\textbf{0.928}$ \color{gray}$_{ \pm 0.002}$ \\
                                                                                            
        Stable Diffusion v3.5 Large \cite{sd3_5_dummy}                                                 & $0.871$ \color{gray}$_{ \pm 0.003}$ 	& $0.864$ \color{gray}$_{ \pm 0.004} $
                                                                                            & $0.876$ \color{gray}$_{ \pm 0.006}$	& $0.866$ \color{gray}$_{ \pm 0.007}$
                                                                                            & $\textbf{0.888}$ \color{gray}$_{ \pm 0.003}$ 	& $\textbf{0.880}$ \color{gray}$_{ \pm 0.003}$ \\

        Flux-Schnell \cite{flux}                                                 & $0.901$ \color{gray}$_{ \pm 0.003}$ 	& $0.892$ \color{gray}$_{ \pm 0.004} $
                                                                                            & $0.915$ \color{gray}$_{ \pm 0.006}$	& $0.901$ \color{gray}$_{ \pm 0.007}$
                                                                                            & $\textbf{0.914}$ \color{gray}$_{ \pm 0.003}$ 	& $\textbf{0.902}$ \color{gray}$_{ \pm 0.003}$ \\

    RAG \cite{flux}                                                 & $0.822$ \color{gray}$_{ \pm 0.003}$ 	& $0.813$ \color{gray}$_{ \pm 0.004} $
                                                                                            & $0.867$ \color{gray}$_{ \pm 0.006}$	& $0.852$ \color{gray}$_{ \pm 0.007}$
                                                                                            & $\textbf{0.853}$ \color{gray}$_{ \pm 0.003}$ 	& $\textbf{0.834}$ \color{gray}$_{ \pm 0.003}$ \\
                                                                                
    \bottomrule
    \end{tabular}
    }
    \caption{GPT-QA scores on T2I Comp-Bench presented in Fig~\ref{fig:flickr_bench_combined}, \algo\textsubscript{X} refers to using \algo\ with X editing model}
    \label{table:t2i_compbench_bench_supp}
\end{table*}

\begin{table*}[h!]
    \label{table:flickr_bench}
    \setlength{\tabcolsep}{6pt}
    \renewcommand{\arraystretch}{1.2}

    \centering
    \resizebox{\textwidth}{!}{
    \begin{tabular}{l|cc|cc|cc}
        \toprule
        \multirow{2}{*}{Method} & \multicolumn{2}{c|}{Base} & \multicolumn{2}{c|}{\algo\textsubscript{AURORA}} & \multicolumn{2}{c}{\algo\textsubscript{PixEdit}}\\
        \cline{2-7}
        & \multicolumn{1}{c}{DSG (w/o dep)} & \multicolumn{1}{c|}{DSG}  & \multicolumn{1}{c}{DSG (w/o dep)} & \multicolumn{1}{c|}{DSG}  & \multicolumn{1}{c}{DSG (w/o dep)} & \multicolumn{1}{c}{DSG} \\
        \midrule

        Stable-Diffusion v1.5 \cite{Rombach_2022_CVPR}                                      & $0.703$ \color{gray}$_{ \pm 0.004}$ 	& $0.640$ \color{gray}$_{ \pm 0.005}$ 
                                                                                            & $\textbf{0.800}$ \color{gray}$_{ \pm 0.016}$ 	& $\textbf{0.748}$ \color{gray}$_{ \pm 0.017}$    
                                                                                            & $0.787$ \color{gray}$_{ \pm 0.005}$ 	& $0.742$ \color{gray}$_{ \pm 0.006}$ \\
                                                                                            
        Structure Diffusion \cite{feng2023trainingfreestructureddiffusionguidance}          & $0.725$ \color{gray}$_{ \pm 0.012}$ 	& $0.648$ \color{gray}$_{ \pm 0.014}$ 
                                                                                            & $\textbf{0.816}$ \color{gray}$_{ \pm 0.003}$ 	& $\textbf{0.766}$ \color{gray}$_{ \pm 0.002}$ 
                                                                                            & $0.802$ \color{gray}$_{ \pm 0.005}$ 	& $0.732$ \color{gray}$_{ \pm 0.006}$ \\
                                                                                            
        Stable Diffusion v2.1 \cite{Rombach_2022_CVPR}                                      & $0.720$ \color{gray}$_{ \pm 0.009}$ 	& $0.656$ \color{gray}$_{ \pm 0.007}$ 
                                                                                            & $0.800$ \color{gray}$_{ \pm 0.010}$ 	& $0.740$ \color{gray}$_{ \pm 0.010}$     
                                                                                            & $\textbf{0.821}$ \color{gray}$_{ \pm 0.001}$ 	& $\textbf{0.757}$ \color{gray}$_{ \pm 0.001}$ \\
                                                                                            
        LMD \cite{lian2024llmgroundeddiffusionenhancingprompt}                              & $0.682$ \color{gray}$_{ \pm 0.006}$ 	& $0.640$ \color{gray}$_{ \pm 0.012}$ 
                                                                                            & $\textbf{0.752}$ \color{gray}$_{ \pm 0.008}$ 	& $\textbf{0.697}$ \color{gray}$_{ \pm 0.008}$ 
                                                                                            & $0.744$ \color{gray}$_{ \pm 0.002}$ 	& $\textbf{0.697}$ \color{gray}$_{ \pm 0.004}$ \\
                                                                                            
        SD-XL \cite{podell2023sdxlimprovinglatentdiffusion}                                 & $0.695$ \color{gray}$_{ \pm 0.004}$ 	& $0.613$ \color{gray}$_{ \pm 0.003}$ 
                                                                                            & $0.775$ \color{gray}$_{ \pm 0.010}$ 	& $0.720$ \color{gray}$_{ \pm 0.011}$        
                                                                                            & $\textbf{0.818}$ \color{gray}$_{ \pm 0.005}$ 	& $\textbf{0.758}$ \color{gray}$_{ \pm 0.007}$ \\
                                                                                            
        PixArt-$\alpha$ \cite{chen2023pixartalphafasttrainingdiffusion}                     & $0.735$ \color{gray}$_{ \pm 0.006}$ 	& $0.663$ \color{gray}$_{ \pm 0.005}$ 
                                                                                            & $0.792$ \color{gray}$_{ \pm 0.005}$ 	& $0.733$ \color{gray}$_{ \pm 0.005}$               
                                                                                            & $\textbf{0.811}$ \color{gray}$_{ \pm 0.011}$ 	& $\textbf{0.767}$ \color{gray}$_{ \pm 0.011}$ \\
                                                                                            
        DeepFloyd IF \cite{DeepFloydIF}                                                     & $0.831$ \color{gray}$_{ \pm 0.003}$ 	& $0.792$ \color{gray}$_{ \pm 0.003}$ 
                                                                                            & $0.842$ \color{gray}$_{ \pm 0.006}$ 	& $0.803$ \color{gray}$_{ \pm 0.007}$               
                                                                                            & $\textbf{0.877}$ \color{gray}$_{ \pm 0.003}$ 	& $\textbf{0.843}$ \color{gray}$_{ \pm 0.005}$ \\
                                                                                            
        PlaygroundV2.5 \cite{li2024playgroundv25insightsenhancing}                          & $0.789$ \color{gray}$_{ \pm 0.005}$ 	& $0.735$ \color{gray}$_{ \pm 0.007}$ 
                                                                                            & $0.806$ \color{gray}$_{ \pm 0.010}$ 	& $0.759$ \color{gray}$_{ \pm 0.012}$            
                                                                                            & $\textbf{0.833}$ \color{gray}$_{ \pm 0.005}$ 	& $\textbf{0.789}$ \color{gray}$_{ \pm 0.004}$ \\
                                                                                            
        Dalle3 \cite{betker2023improving}                                                   & $0.891$ \color{gray}$_{ \pm 0.004}$ 	& $0.857$ \color{gray}$_{ \pm 0.005}$ 
                                                                                            & $0.890$ \color{gray}$_{ \pm 0.002}$ 	& $0.847$ \color{gray}$_{ \pm 0.001}$              
                                                                                            & $\textbf{0.904}$ \color{gray}$_{ \pm 0.004}$ 	& $\textbf{0.873}$ \color{gray}$_{ \pm 0.007}$ \\
                                                                                            
        Stable Diffusion v3.5 Large \cite{sd3_5_dummy}                                                 & $0.881$ \color{gray}$_{ \pm 0.004}$ 	& $0.852$ \color{gray}$_{ \pm 0.004}$ 
                                                                                            & $0.905$ \color{gray}$_{ \pm 0.002}$ 	& $0.873$ \color{gray}$_{ \pm 0.004}$    
                                                                                            & $\textbf{0.909}$ \color{gray}$_{ \pm 0.003}$ 	& $\textbf{0.884}$ \color{gray}$_{ \pm 0.003}$ \\

    Flux-Schnell \cite{flux}                                                 & $0.884$ \color{gray}$_{ \pm 0.003}$ 	& $0.864$ \color{gray}$_{ \pm 0.004} $
                                                                                            & $0.876$ \color{gray}$_{ \pm 0.003}$	& $0.846$ \color{gray}$_{ \pm 0.004}$
                                                                                            & $\textbf{0.890}$ \color{gray}$_{ \pm 0.001}$ 	& $\textbf{0.871}$ \color{gray}$_{ \pm 0.002}$ \\

    RAG \cite{flux}                                                & $0.750$ \color{gray}$_{ \pm 0.003}$ 	
                                                                                    & $0.676$ \color{gray}$_{ \pm 0.004} $
                                                                                    & $0.809$ \color{gray}$_{ \pm 0.003}$	& $0.756$ \color{gray}$_{ \pm 0.004}$
                                                                                    & $\textbf{0.816}$ \color{gray}$_{ \pm 0.002}$ 	& $\textbf{0.782}$ \color{gray}$_{ \pm 0.002}$ \\                                                                                     
    \bottomrule
    \end{tabular}
    }
    \caption{GPT-QA scores on Flick-Bench presented in Fig~\ref{fig:flickr_bench_combined}, \algo\textsubscript{X} refers to using \algo\ with X editing model }
    \label{tab:flickr_bench_detailed}
\end{table*}

\begin{figure*}
  \centering
  \includegraphics[width=1.0\linewidth]{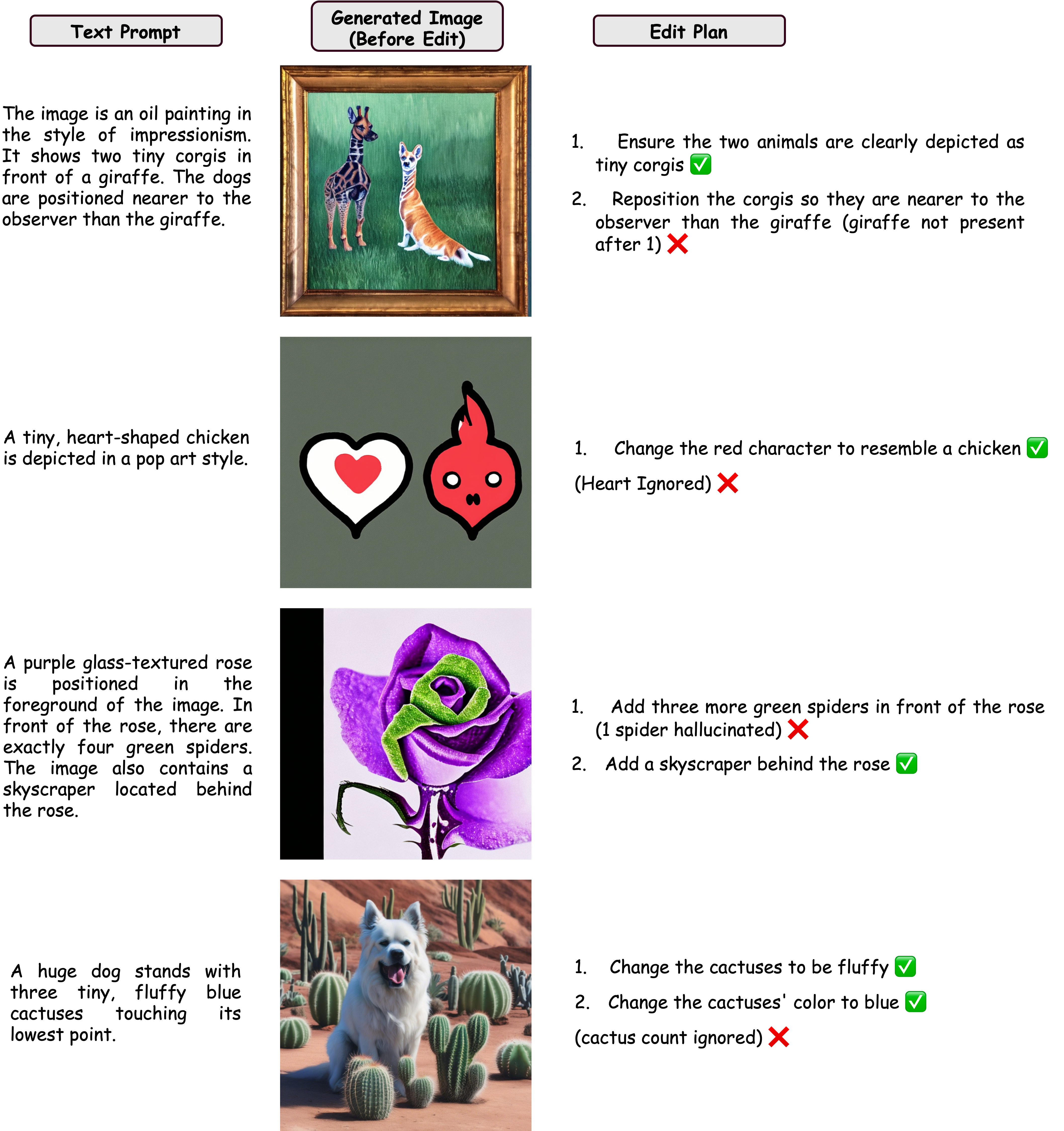}
  \caption{Results illustrating failure cases of generated Edit Plans}
  \label{fig:plan_failures}
\end{figure*}

\begin{figure*}
  \centering
  \includegraphics[width=1.0\linewidth]{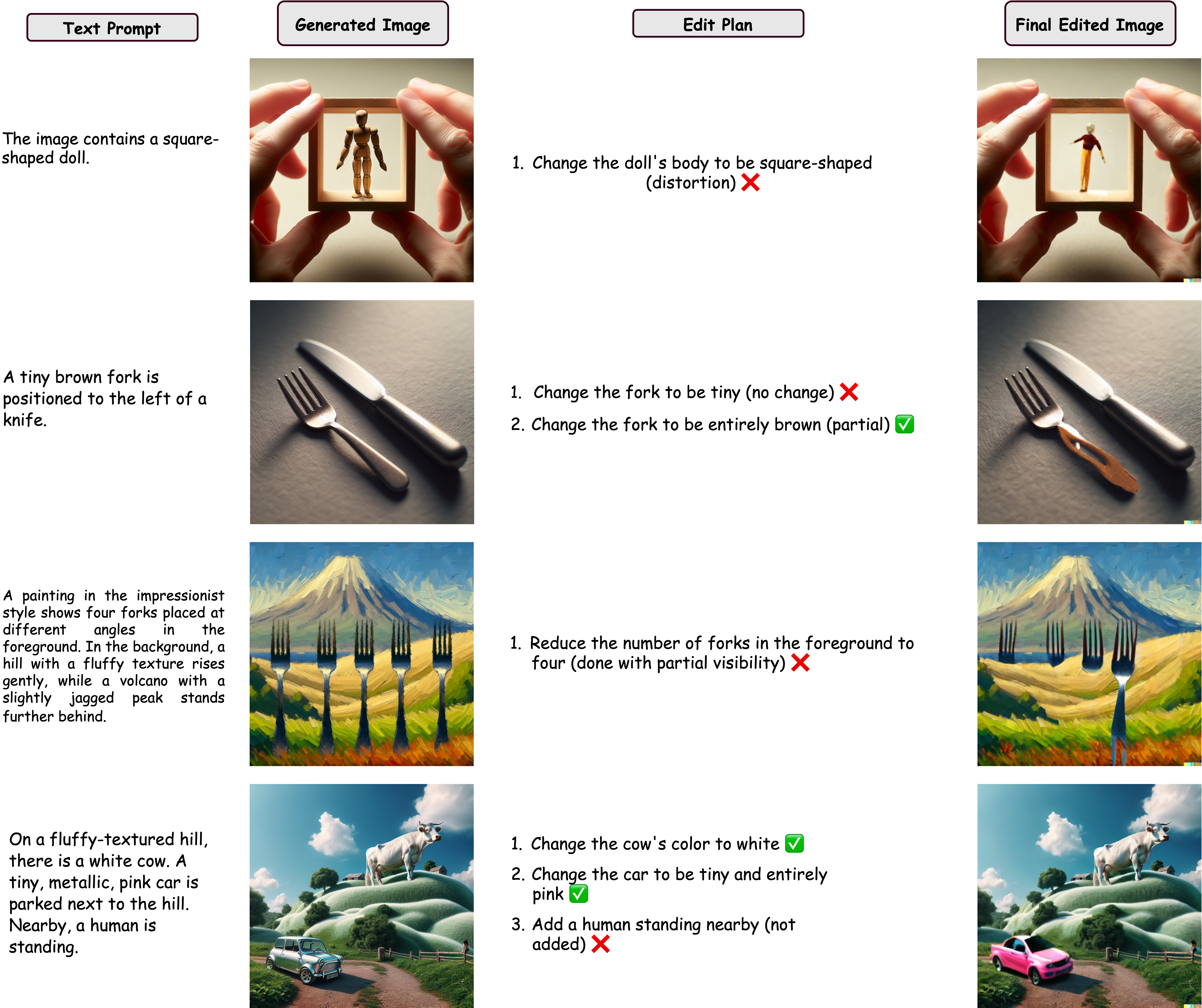}
  \caption{More results illustrating failure cases of Editing Models}
  \label{fig:edit_failures}
\end{figure*}
